\documentclass[11pt, a4paper, logo, copyright, nonumbering]{main}
\usepackage[authoryear, sort&compress, round]{natbib}
\usepackage{dblfloatfix}
\usepackage{ulem}
\usepackage{caption}
\usepackage{dramatist}
\usepackage{xspace}
\usepackage{pifont} %
\usepackage{multirow}
\usepackage{tcolorbox}
\usepackage{xltabular}
\usepackage{longtable}
\usepackage{hyperref}
\usepackage{amsfonts}
\usepackage{amsmath}
\usepackage{amssymb}
\usepackage{lineno}
\usepackage{multirow}
\usepackage{adjustbox}

\usepackage[bottom]{footmisc}

\usepackage{CJKutf8}
\usepackage{subfigure}
\usepackage{setspace}

\usepackage{dsfont}
\usepackage{array} %
\usepackage{tabularx} %
\usepackage{subfigure} %
\usepackage{xcolor} %
\usepackage{tabularx}
\usepackage{booktabs}
\usepackage{xspace}

\usepackage{lipsum}  %
\usepackage{multicol} %

\usepackage{cleveref}
\usepackage{epigraph}

\makeatletter
\def\@BTrule[#1]{%
  \ifx\longtable\undefined
    \let\@BTswitch\@BTnormal
  \else\ifx\hline\LT@hline
    \nobreak
    \let\@BTswitch\@BLTrule
  \else
    \let\@BTswitch\@BTnormal
  \fi\fi
  \global\@thisrulewidth=#1\relax
  \ifnum\@thisruleclass=\tw@\vskip\@aboverulesep\else
  \ifnum\@lastruleclass=\z@\vskip\@aboverulesep\else
  \ifnum\@lastruleclass=\@ne\vskip\doublerulesep\fi\fi\fi
  \@BTswitch
}
\makeatother

\addto\extrasenglish{
}

{\begin{list}{}%
        {\setlength{\leftmargin}{#1}}%
        \item[]%
}
{\end{list}}

\reportnumber{001} %

\renewcommand{\today}{}

\def\Ours{\text{NextStep-1}\xspace}
\def\Oursedit{\text{NextStep-1-Edit}\xspace}

\title{\centering NextStep-1: Toward Autoregressive Image Generation with Continuous Tokens at Scale}

\def\huggingface{\raisebox{-1.5pt}{\includegraphics[height=1.05em]{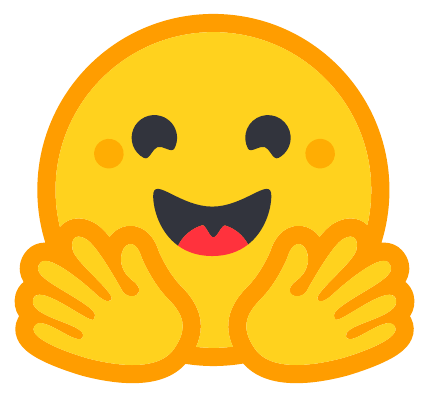}}}
\def\github{\raisebox{-1.5pt}{\includegraphics[height=1.05em]{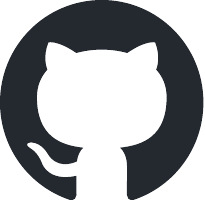}}}
\def\homepage{\raisebox{-1.5pt}{\includegraphics[height=1.2em]{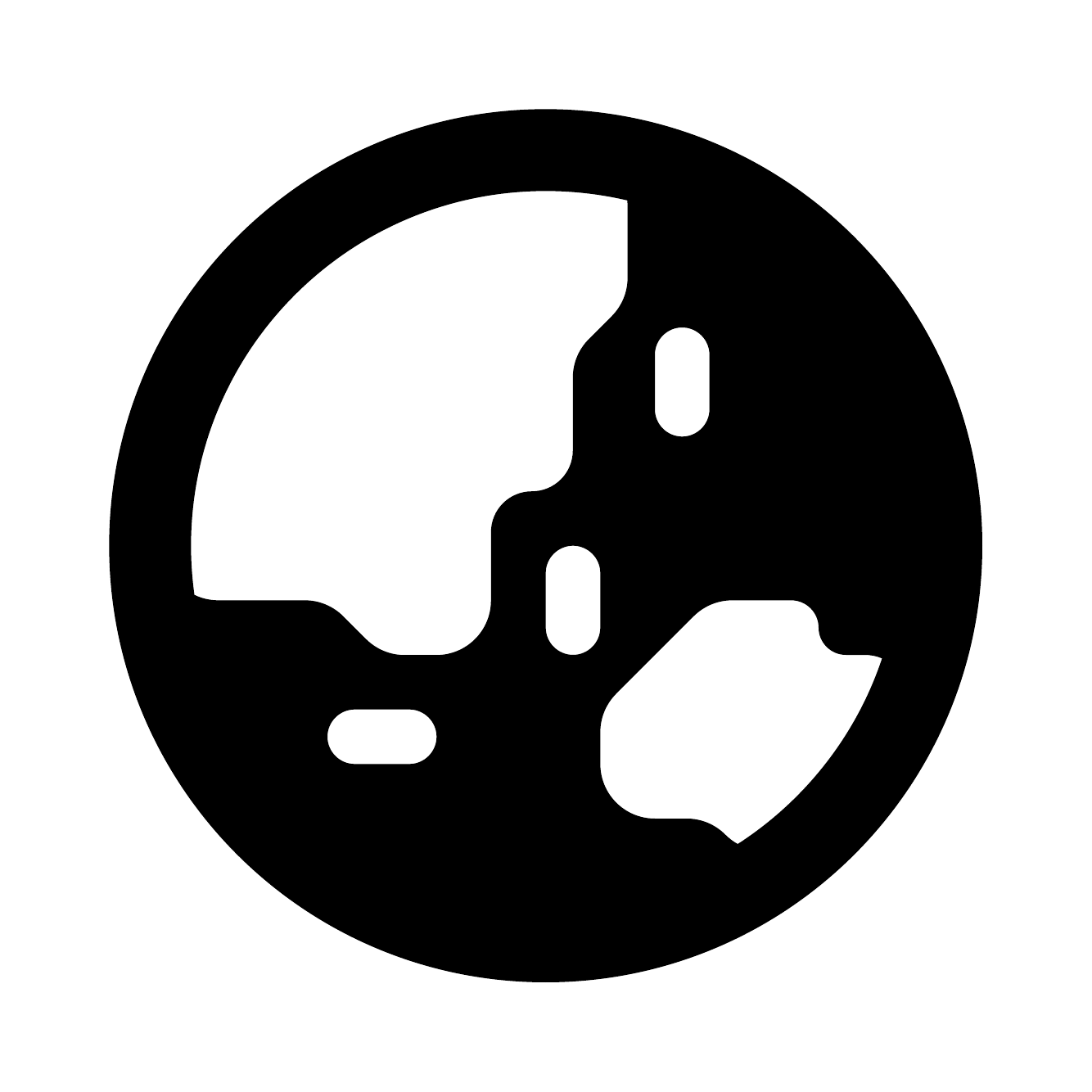}}}

\author[*]{
\textbf{NextStep-Team, StepFun}
\\
\vspace{-0.7em}
\small \homepage~\textbf{Homepage}: \url{https://stepfun.ai/research/en/nextstep1} \\
\small \github~\textbf{Github}: \url{https://github.com/stepfun-ai/NextStep-1} \\
\small \huggingface~\textbf{Huggingface}: \href{https://huggingface.co/collections/stepfun-ai/nextstep-1-689d80238a01322b93b8a3dc}{NextStep-1 Collections}
}

\renewcommand{\phi}{\varphi}

\renewcommand{\epsilon}{\varepsilon}
\renewcommand{\imath}{\mathrm{i}}

\newlength{\restsubwidth}
\newlength{\restsubheight}
\newlength{\restsubmoreheight}
\newcommand{\rest}[2]{%
        \settowidth{\restsubwidth}{\ensuremath{#2}}
        \settoheight{\restsubheight}{\ensuremath{{}_{#2}}}
        \ensuremath{{#1\hskip 0.5pt}_{\vrule\kern2pt\parbox[b][%
        4pt][b]{\the\restsubwidth}{%
                        \ensuremath{{}_{#2}}}}}
        }

\begin{abstract}

Prevailing autoregressive (AR) models for text-to-image generation either rely on heavy, computationally-intensive diffusion models to process continuous image tokens, or employ vector quantization (VQ) to obtain discrete tokens with quantization loss. In this paper, we push the autoregressive paradigm forward with \textbf{\Ours}, a 14B autoregressive model paired with a 157M flow matching head, training on discrete text tokens and continuous image tokens with next-token prediction objectives. \Ours achieves state-of-the-art performance for autoregressive models in text-to-image generation tasks, exhibiting strong capabilities in high-fidelity image synthesis. Furthermore, our method shows strong performance in image editing, highlighting the power and versatility of our unified approach. To facilitate open research, we will release our code and models to the community.

\end{abstract}

\begin{document}

\maketitle

%%%%%%%%%%%%%%%%%%%%%%%
%% color definition %%%

\definecolor{colorfirst}{RGB}{252,141,89}
\definecolor{colorsecond}{RGB}{253,187,132}
\definecolor{colorthird}{RGB}{253,212,158}
\definecolor{colorfourth}{RGB}{254,232,200}
\definecolor{colorfifth}{RGB}{255,247,236}
\definecolor{myred}{RGB}{242,128,128}
\definecolor{mygreen}{RGB}{112,180,143}
\definecolor{myblue}{RGB}{210,225,255}
\definecolor{citypink}{RGB}{227,108,194}
\definecolor{cityblue}{RGB}{128,159,225}
\newcommand{\rankfirst}[0]{\cellcolor{colorfirst}}
\newcommand{\ranksecond}[0]{\cellcolor{colorsecond}}
\newcommand{\rankthird}[0]{\cellcolor{colorthird}}
\newcommand{\rankfourth}[0]{\cellcolor{colorfourth}}
\newcommand{\rankfifth}[0]{\cellcolor{colorfifth}}
\DeclareRobustCommand{\legendsquare}[1]{%
  \textcolor{#1}{\rule{2ex}{2ex}}%
}
\DeclareRobustCommand{\legendsquarebox}[1]{%
  \tikz[] \draw[black, fill=#1, line width=0.4pt] (0,0) rectangle (1.5ex,1.5ex);%
}
\newcommand{\cmark}{\textcolor{mygreen}{\ding{51}}}%
\newcommand{\xmark}{\textcolor{myred}{\ding{55}}}%

%%%%%%%%%%%%%%%%%%%%%%%

\section{Introduction}

\begin{figure*}[t!]
\centering
\includegraphics[width=.95\linewidth]{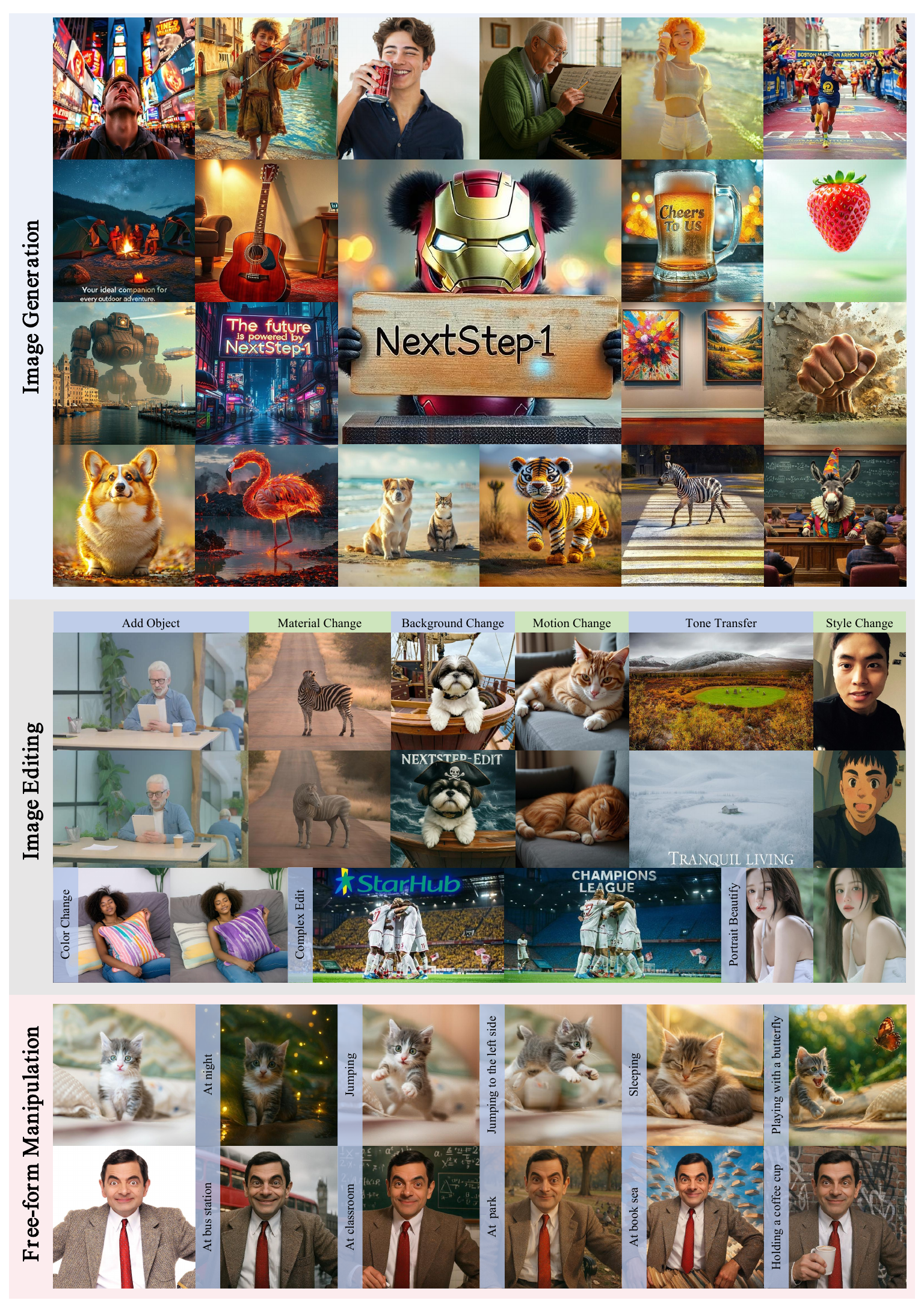}
\caption{
Overview of \Ours in high-fidelity image generation, diverse image editing, and complex free-form manipulation.
}
\label{fig:teaser}
\end{figure*}

The remarkable success of autoregressive models in large language models~\citep{gpt1,gpt2,gpt3,gpt4.1} has motivated their extension to text-to-image generation. By unifying multimodal inputs into a single sequence, autoregressive image generation models~\citep{parti,emu1,emu2,emu3,fan2024fluid,sun2024latentlm, januspro2025} offer a scalable and flexible approach to text-to-image generation that naturally accommodates diverse conditioning signals.

However, most existing autoregressive approaches for text-to-image generation~\citep{emu2, emu3, sun2024llamagen, januspro2025, tong2024metamorph, dreamllm} either rely on heavy diffusion models or adopt vector quantization (VQ)~\citep{movqgan, eslami2021taming, yu2023magvitv2} to tokenize images into discrete visual tokens, which encounter limitations including exposure bias~\citep{han2025infinity} and suboptimal image
tokenization~\citep{mar}. While recent efforts with continuous latent representations~\citep{mar,fan2024fluid, sun2024latentlm,tschannen2025givt, tschannen2024jetformer} have shown promise, a significant performance gap persists between autoregressive models and state-of-the-art diffusion methods~\citep{sdxl,SD3,flux2024}, particularly in image quality and consistency.

In this paper, we introduce \Ours, a simple yet effective autoregressive model built on the next-token prediction paradigm that achieves state-of-the-art performance in text-to-image generation tasks. Comprehensive evaluations confirm its competitive performance across a suite of challenging benchmarks. Specifically, \Ours demonstrates exceptional compositional and linguistic understanding, achieving \textbf{0.54} on WISE~\citep{niu2025wise}, \textbf{0.67} on the advanced prompts of GenAI-Bench~\citep{lin2024genaibench}, \textbf{85.28} on DPG-Bench~\citep{hu2024ella}, and \textbf{0.417} on the English prompts of OneIG-Bench~\citep{chang2025oneigbench}. These results demonstrate its capabilities across diverse scenarios, from short and long prompts to tasks requiring world knowledge. Beyond generation, the versatility of \Ours is validated by its strong performance on instruction-based image editing, \Oursedit, achieving competitive scores of \textbf{6.58} for English prompts on GEdit-Bench~\citep{liu2025step1x} and \textbf{3.71} on ImgEdit-Bench~\citep{ye2025imgedit}. We showcase the qualitatve performance in \Cref{fig:teaser}.

\Ours is a 14-billion-parameter autoregressive model composed of a Transformer backbone, a standard language modeling head for discrete text tokens, a lightweight flow matching head for continuous image tokens, and an image tokenizer. The flow matching head is a 157-million-parameter, MLP-based model trained with a flow matching objective, following the approach of \citep{mar}. In autoregressive modeling, high-dimensional latent spaces are critical for achieving high image quality but often induce training instability and divergence. Our image tokenizer addresses this trade-off by enhancing the robustness of continuous image tokens and promoting a well-dispersed, normalized latent space, thereby ensuring stable convergence even at higher dimensionalities (e.g., 16 channels). Empirical results confirm that this design is essential for stable and effective training with 16-channel latents.

\section{Framework}

\begin{figure*}[t!]
\centering
\includegraphics[width=\linewidth]{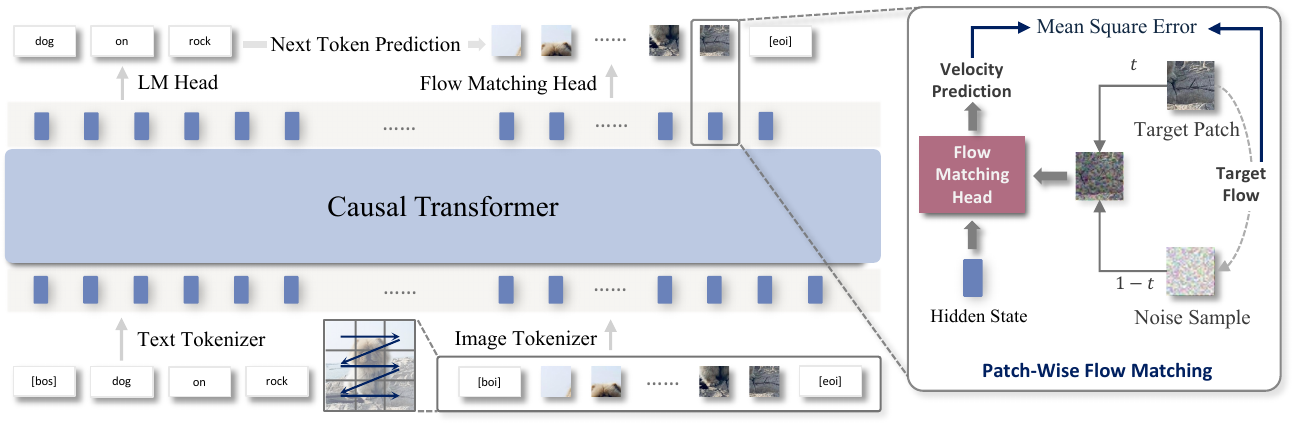}
\caption{\small
% \textbf{Overview of our \Ours~framework}.
% Image patches are tokenized and flattened in raster-scan order. During training, patch-wise flow matching loss supervises transitions between tokens. At inference, tokens are generated from noise, guided by hidden states via the Flow Matching Head.
Overview of \Ours~Framework. \Ours~employs a causal transformer to process tokenized text and image tokens. During training, Flow Matching Head predicts the continuous flow from a noise sample to the next target image patch, conditioned on the output hidden state. At inference, this allows for generating images by iteratively guiding noise to create the next patch.
% \textbf{Overview of \Ours~Framework.} \Ours~employs a causal transformer to process tokenized text and image patches. Flow Matching Head is trained to predict the continuous flow from a noise sample to the next target image patch, conditioned on the output hidden state. At inference, this allows for generating images by iteratively guiding noise to create the next patch.
}
\label{fig:model_arch}
\end{figure*}

\subsection{Unified Multi-model Generation with Continuous Visual Tokens}

\Ours extends the well-established autoregressive language modeling paradigm to image generation through a simple and intuitive architecture, as illustrated in \Cref{fig:model_arch}. To unify multimodal inputs into a single sequence, the images will be tokenized to \textbf{continuous image tokens} by the image tokenizer and combined with discrete text tokens. Supposing $x = \{x_{0}, x_{1}, ..., x_{n}\}$ is the multimodel token sequence, where $x_i$ is either a discrete text token or a continuous visual token, the autoregressive objective under the unified sequence is formalized as:
\begin{equation}
p(x) = \prod_{i=1}^{n} p(x_i \mid x_{<i}).
\end{equation}
The unified multi-modal generation task proceeds by sampling the next token \(x_{i}\) from the conditional distribution \(p(x_i \mid x_{<i})\) modeled by a network. Discrete text tokens are sampled via a language modeling head, while continuous image tokens are sampled by a flow-matching head.

Our training objective consists of two distinct losses: a standard cross-entropy loss for discrete text tokens, and a flow matching loss~\citep{ipman2023flowmatching} for continuous image tokens. Specifically, the flow matching loss is the mean squared error between the predicted and target velocity vectors that map a noised patch to its corresponding clean patch. The model is trained end-to-end by optimizing a weighted sum of these two losses:
\begin{equation}
\mathcal{L}_{\text{total}} = \lambda_\text{text}\mathcal{L}_{\text{text}} + \lambda_\text{visual} \mathcal{L}_{\text{visual}}
\label{eq:loss_total}
\end{equation}
where $\mathcal{L}_{\text{text}}$ and $\mathcal{L}_{\text{visual}}$ denote the loss for text and image tokens respectively, which are balanced by the hyperparameters $\lambda_\text{text}$ and $\lambda_\text{visual}$.

\subsection{Model Architecture}

\paragraph{Image Tokenizer.}  Our image tokenizer is fine-tuned from flux VAE~\citep{flux2024} with only reconstruction and perceptual losses. The tokenizer first encodes an image into 16-channel latents $z$, applying an 8$\times$ spatial downsampling factor. To stabilize and normalize the latent space, we apply channel-wise normalization, standardizing each channel to zero mean and unit variance. Furthermore, to enhance the robustness of the image tokenizer and encourage a more uniform latent distribution, we introduce a stochastic perturbation to the normalized latents. This technique is adapted from $\sigma$-VAE~\citep{sun2024latentlm}, where it was employed to prevent variance collapse.
\begin{equation}
\tilde{z}=\text{Normlization}(z)+\alpha \cdot \epsilon, \quad \text{where}~ \alpha \sim \mathcal{U}[0,~\gamma] ~\text{and}~ \epsilon \sim \mathcal{N}(0,~I)
\label{eq:add_noise}
\end{equation}
where $\epsilon$ is standard Gaussian noise, and its magnitude is scaled by a random factor $\alpha$ sampled uniformly from [0,~$\gamma$]. The $\gamma$ is a hyperparameter controlling the maximum noise intensity.
 
The latents from the image tokenizer are pixel-shuffled into a more compact sequence. This is achieved by applying a space-to-depth transformation with a 2$\times$2 kernel, which flattens 2$\times$2 spatial latents into the channel dimension. For example, this converts the latents of a 256$\times$256 image into a 16$\times$16 grid of 64-channel tokens. This grid is then flattened into a 1D sequence of 256 tokens to serve as input for the following Causal Transformer.

\paragraph{Causal Transformer.} We initialize our model from the decoder-only Qwen2.5-14B~\citep{qwen2.5}, leveraging its strong language understanding and reasoning capabilities for text-to-image generation. We organize the multimodal input sequence in the following format:
\[
\small{\text{\{text\} <image\_area>h*w <boi> \{image\} <eoi>}}...
\]
where $\text{\{text\}}$ denotes discrete text tokens, and $\text{\{image\}}$ represents continuous image tokens. \text{<boi>} and \text{<eoi>} are special tokens marking the beginning-of-image and end-of-image. $\text{<image\_area>h*w}$ represents the metadata about the spatial dimensions of the 2D image tokens.

Then the output hidden states from LLM are passed to two lightweight heads for modality-specific loss:
\begin{itemize}
\item \textbf{Language Modeling Head.} We compute Cross-Entropy loss for hidden states of texts. 
\item \textbf{Patch-wise Flow Matching Head.} Following ~\citep{mar}, we use each patch-wise image hidden states as condition, denoise target patch at timesteps t, and compute the patch-wise flow-matching loss~\citep{lipman2022flow} with a 157M, 12-layer, and 1536 hidden-dimensions MLP.
\end{itemize}

For positional information, we use the standard 1D RoPE~\citep{su2024roformer}. Despite the availability of more complex 2D or multimodal RoPE alternatives~\citep{qwen2.5-vl,qwen2vl}, we found that the simple 1D formulation remains highly effective for mixed text-image sequences, and thus retain it for simplicity and efficiency.
\section{Data}
\label{sec:data}

To comprehensively equip our model with broad and versatile capabilities, we construct a diverse training corpus composed of four primary data categories: a text-only corpus, image-text pair data, image-to-image data, and interleaved data. Each category is curated to serve a distinct role in fostering different aspects of the model's generative abilities.

\subsection{Text-only Corpus} 

To preserve the extensive language capabilities inherent in the large language model (LLM), we incorporate \textbf{400B text-only tokens} sampled from Step-3~\citep{stepfun2025step3largeaffordablemodelsystem} during training.

\subsection{Image-Text Pair Data}
\label{sec:image_text_pair}

Data consisting of image-text pairs forms the foundation of the model's text-to-image generation capabilities. We developed a comprehensive pipeline to curate a high-quality, large-scale dataset from a diverse set of initial sources.
\begin{enumerate}
    \item Data Sourcing: We collected a large-scale dataset from diverse sources, including web data, multi-task VQA data, and text-rich documents.
    \item Quality-Based Filtering: We then applied a rigorous filtering process, evaluating each image on aesthetic quality, watermark presence, clarity, OCR detection, and text-image semantic alignment.
    \item Re-captioning: After deduplicating the filtered images, we used the Step-1o-turbo \footnote{\url{https://platform.stepfun.com/docs/llm/vision}} to generate rich and detailed captions for each image in both English and Chinese.
\end{enumerate}
This multi-stage pipeline yields a final dataset of \textbf{550M high-quality image-text pairs}, providing a foundation for training a model with both strong aesthetic sense and broad world knowledge.

\subsection{Instruction-Guided Image-to-Image Data}
\label{sec:image-to-image-data}
To enable a wide range of practical applications, we curated a high-quality dataset for instruction-guided image-to-image tasks, such as visual perception~\citep{kirillov2023segment}, controllable image generation~\citep{zhang2023controlnet}, image restoration~\citep{fluxfill}, general image editing~\citep{peng2024dreambench++}, and more.

For visual perception and controllable image generation tasks, we synthesized 1M samples by applying the annotator of ControlNet~\citep{zhang2023controlnet} to a part of our high-quality image-text pair data. For image restoration and
general image editing, we collected 3.5M samples, comprising data from GPT-Image-Edit~\citep{wang2025gpt-edit-1_5M}, Step1X-Edit~\citep{liu2025step1x}, and a proprietary in-house dataset. Following Step1X-Edit~\citep{liu2025step1x}, all editing data were subjected to a rigorous VLM-based filtering pipeline that assessed both image-pair quality, rationality, consistency, and instruction alignment, resulting in about \textbf{1M high-quality instruction-guided image-to-image data} for training.

\subsection{Interleaved data} 

Interleaved data seamlessly integrates text and images, offering rich and nuanced sequential associations between modalities. Specifically, our knowledge-rich interleaved dataset is primarily composed of four distinct categories: general video-interleaved data, tutorials, character-centric scenes, and multi-view data. 

\begin{figure}
    \centering
    \includegraphics[width=1.0\linewidth]{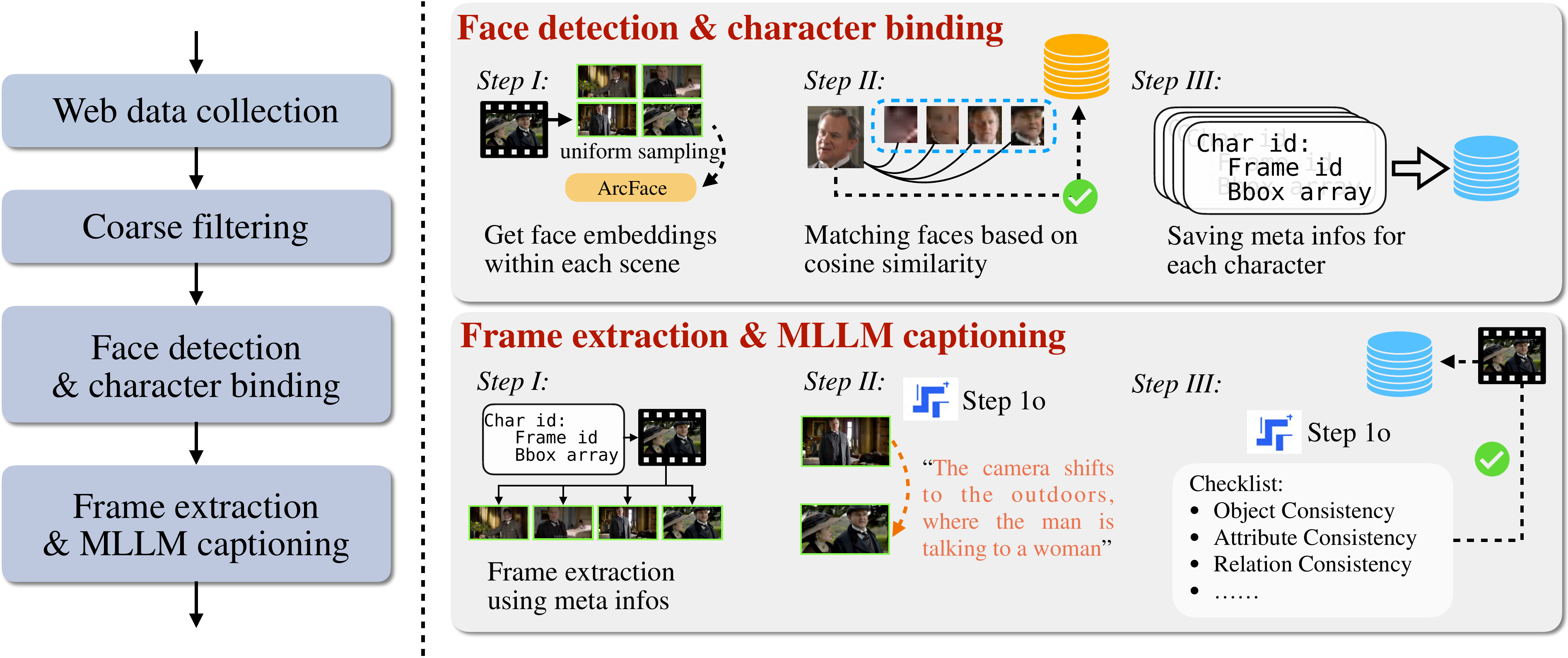}
    \caption{Data processing of character-centric data.}
    \label{fig:char_data}
\end{figure}

To endow our model with extensive world knowledge, we first constructed a large-scale, 80M-sample video-interleaved dataset. This was achieved through a meticulous curation pipeline, inspired by Step-Video~\citep{Step-Video-T2V}, which encompasses frame extraction, deduplication, and captioning. Furthermore, following the methodology of mmtextbook~\citep{mmtextbook}, we collected and processed tutorial videos by leveraging ASR and OCR tools. This component specifically targets text-rich real-world scenes, enhancing the model's textual understanding and generation in context. A key contribution, detailed in \Cref{fig:char_data}, is our character-centric dataset, \textbf{NextStep-Video-Interleave-5M}. For this dataset, we extracted video frames centered around specific characters and generated rich, storytelling-style captions akin to~\citep{oliveira2025storyreasoning}, thereby significantly improving the model's capacity for multi-turn interaction. Finally, to bolster geometric reasoning, we curated multiview data from two open-source datasets, MV-ImageNet-v2~\citep{han2024mvimgnet2} and Objaverse-XL~\citep{deitke2023objaverse}, which enhances the model's ability to maintain multiview consistency.

\section{Training Recipe}

\subsection{Training Image Tokenizer}

Our image tokenizer is initialized from the Flux.1-dev VAE~\citep{flux2024}, selected for its strong reconstruction performance. We fine-tune this model on the image-text dataset detailed in \Cref{sec:image_text_pair} to adapt it to our specific data distribution. For optimization, we employ the AdamW optimizer~\citep{loshchilov2019adamw} with $(\beta_1= 0.9, \beta_2 = 0.95, \epsilon=1 \times 10^{-8})$ for its convergence stability. The model is trained for 50K steps with a total batch size of 512, using a constant learning rate of $1 \times 10^{-5}$ preceded by a linear warm-up of 1000 steps.

\begin{table}[ht]
\centering
\scriptsize
\caption{\small Training recipe of \Ours.}
% \vspace{0.2cm}
\resizebox{\columnwidth}{!}{
\begin{tabular}{l|ccccc}
\toprule
& \multicolumn{3}{c}{\textbf{Pre-Training}} & \multicolumn{2}{c}{\textbf{Post-Training}} \\
\cmidrule(lr){2-4}
\cmidrule(lr){5-6}
& \textbf{Stage1} & \textbf{Stage2} & \textbf{Annealing} & \textbf{SFT} & \textbf{DPO} \\
\midrule
\textbf{Hyperparameters} & & & & & \\
\midrule
Learning Rate (Min, Max) & $1 \times 10^{-4}$ & $1 \times 10^{-5}$ & (0, $1 \times 10^{-5}$) & (0, $1 \times 10^{-5}$) & $2 \times 10^{-6}$ \\
LR Scheduler & Constant & Constant & Cosine & Cosine & Constant \\
Weight Decay & 0.1 & 0.1 & 0.1 & 0.1 & 0.1 \\
Loss Weight (CE : MSE) & (0.01 : 1) & (0.01 : 1) & (0.01 : 1) & (0.01 : 1) & - \\
Training Steps & 200K & 100K & 20K & 10K & 300 \\
Warm-up Steps & 5K & 5K & 0 & 500 & 200 \\
Sequence Length per Rank & 16K & 16K & 16K & 8K & - \\
Image Area (Min, Max) & 256$\times$256 & (256$\times$256, 512$\times$512) & (256$\times$256, 512$\times$512) & (256$\times$256, 512$\times$512) & (256$\times$256, 512$\times$512) \\
Image Tokens (Min, Max) & 256 & (256, 1024) & (256, 1024) & (256, 1024) & (256, 1024) \\
Training Tokens & 1.23T & 0.61T & 40B & 5B & - \\
\midrule
\textbf{Data Ratio} & & & & & \\
\midrule
Text-only Corpus & 0.2 & 0.2 & 0.2 & 0 & - \\
Image-Text Pair Data & 0.6 & 0.6 & 0.6 & 0.9 & - \\
Image-to-Image Data & 0.0 & 0.0 & 0.1 & 0.1 & - \\
Interleaved Data & 0.2 & 0.2 & 0.1 & 0 & - \\
\bottomrule
\end{tabular}
}
\label{table:training_recipe}
\end{table}
\subsection{Pre-Training}

The specific hyperparameters and data ratios for our pre-training are detailed in \Cref{table:training_recipe}.
Specifically, the pre-training follows a three-stage curriculum designed to progressively refine the model's capabilities. Throughout these stages, all model parameters are trained end-to-end except for the pre-trained image tokenizer.

\paragraph{Stage1.} In this initial stage, the model learns a foundational understanding of image structure and composition. For computational efficiency, all images are resized and randomly cropped to a fixed 256$\times$256 resolution. The training curriculum is composed of a diverse data mixture: 20\% text-only corpora, 60\% image-text pairs, and 20\% interleaved data. This stage consumed approximately 1.23T tokens.

\paragraph{Stage2.} We implement a dynamic resolution strategy to train the model on a range of higher resolutions, targeting 256$\times$256 and 512$\times$512 base areas. This strategy utilizes different aspect ratio buckets for computational efficiency. In this stage, we enrich the data mixture with more text-rich and video-interleaved data, leveraging the model's enhanced capacity to process fine details at these resolutions.

\paragraph{Annealing.} In the final stage of pre-training, we perform an annealing phase to sharpen the model's capabilities on a highly curated dataset. This is achieved by training the model for one epoch on a high-quality subset of 20M samples, which were selected from \Cref{sec:image_text_pair} by applying stricter filtering thresholds for aesthetic score, image clarity, semantic similarity, watermark, and so on. This annealing step significantly improves the model's final output, enhancing overall image structure, composition, texture, and aesthetic appeal.

\subsection{Post-Training}
Following pre-training on a broad corpus to establish a generalist model, post-training serves to align the model's output with human preferences and downstream tasks. We achieve this alignment via a two-stage process: Supervised Fine-Tuning (SFT) followed by Direct Preference Optimization (DPO)~\citep{rafailov2023direct}. The hyperparameters for each stage are in \Cref{table:training_recipe}.

\paragraph{Supervised Fine-Tuning (SFT).} The SFT stage enhances the model’s instruction-following capabilities and aligns its outputs with human preferences. The SFT dataset, comprising a total of 5M samples, is organized into three components: 1) a corpus of human-selected image-text pairs with high semantic consistency and visual appeal, augmented by images from other generative models to improve the model's handling of complex and imaginative prompts through distillation; 2) Chain-of-Thought (CoT) data~\citep{COT,deng2025emerging}, improving text-to-image generation by incorporating a language-based reasoning step before the final image is created; 3) high-quality instruction-guided image-to-image data from \Cref{sec:image-to-image-data} to strengthen the model's image editing capabilities.

\paragraph{Direct Policy Optimization (DPO).} To align our model with human preferences, we employ Direct Policy Optimization (DPO) ~\citep{rafailov2024dpo}, a method inspired by Diffusion-DPO ~\citep{wallace2024diffusiondpo}. To this end, we construct two distinct types of preference datasets from a curated set of approximately 20,000 diverse prompts.
\begin{enumerate}
\item Standard DPO Dataset: For each prompt $c$, we directly use the SFT model to generate 16 candidate images. These images is then scored by ImageReward~\citep{xu2023imagereward} to form a preference pair $(y^{w}, y^{l})$, where the winning image $y^{w}$ is randomly sampled from the top 4 candidates, while the losing image $y^{l}$ is randomly sampled from the remaining 12. 

\item Self-CoT DPO Dataset: To enhance the model's reasoning capabilities, we introduce an explicit reasoning step. For each prompt $c$, we first prompt our model to generate a detailed textual CoT, which is then extended to the original prompt. Using this CoT-enhanced prompt, we follow the identical pipeline as above to form a preference pair $(y^{w}, y^{l})$.
\end{enumerate}
\begin{table}[t!]
\centering
\scriptsize
\caption{\small Comparison of image-text alignment on GenEval~\citep{ghosh2023geneval}, GenAI-Bench~\citep{lin2024genaibench}, and DPG-Bench~\citep{hu2024ella}. *~result is with rewriting. $\dag$~result is with Self-CoT.}
\resizebox{\columnwidth}{!}{
\begin{tabular}{lcccc}
\toprule
\multirow{2}{*}{\textbf{Method}} & \multirow{2}{*}{\textbf{~~~~GenEval$\uparrow$~~~~}} & \multicolumn{2}{c}{\textbf{GenAI-Bench$\uparrow$}} & \multirow{2}{*}{\textbf{~~~~DPG-Bench$\uparrow$~~~~}} \\
\cmidrule(lr){3-4}
& & \textbf{~~~~~~Basic~~~~~~} & \textbf{~~Advanced~~} & \\
% & & \textbf{Basic} & \textbf{Advanced} & \\
\midrule
\textbf{\textit{Proprietary}} \\
\midrule
DALL-E 3~\citep{dalle3} & 0.67 & \emph{0.90} & 0.70 & 83.50 \\
Seedream 3.0~\citep{gao2025seedream_v3} & 0.84 &- &- & \emph{88.27} \\
GPT4o~\citep{openai2024gpt4o_image} & 0.84 &- &- & 85.15 \\
\midrule
\textbf{\textit{Diffusion}} \\
\midrule
Stable Diffusion 1.5~\citep{sdv15} & 0.43 &- &- &- \\
Stable Diffusion XL~\citep{sdxl} & 0.55 & 0.83 & 0.63 & 74.65 \\
Stable Diffusion 3 Medium~\citep{SD3} & 0.74 & 0.88 & 0.65 & 84.08 \\
Stable Diffusion 3.5 Large~\citep{SD3} & 0.71 & 0.88 & 0.66 & 83.38 \\
PixArt-Alpha~\citep{chen2024pixart} & 0.48 &- &- & 71.11 \\
Flux.1-dev~\citep{flux2024} & 0.66 & 0.86 & 0.65 & 83.79 \\
Transfusion~\citep{transfusion} & 0.63 &- &- &- \\
CogView4~\citep{2025cogview4} & 0.73 &- &- & 85.13 \\
Lumina-Image 2.0~\citep{qin2025lumina2} & 0.73 &- &- & \emph{87.20} \\
HiDream-I1-Full~\citep{cai2025hidream} & 0.83 & \textbf{0.91} & 0.66 & 85.89 \\
Mogao~\citep{liao2025mogao} & \textbf{0.89} &- & 0.68 & 84.33 \\
BAGEL~\citep{deng2025emerging} & 0.82/\emph{0.88}$^\dag$ & 0.89/0.86$^\dag$ & 0.69/\textbf{0.75}$^\dag$ & 85.07 \\
% BAGEL $w/$ Self-CoT~\citep{deng2025emerging} & \emph{0.88} & 0.86 & \textbf{0.75} & \\
Show-o2-7B~\citep{xie2025showo2} & 0.76 &- &- & 86.14 \\
OmniGen2~\citep{wu2025omnigen2} & 0.80/0.86* &- &- & 83.57 \\
Qwen-Image~\citep{wu2025qwen_image} & 0.87 &- &- & \textbf{88.32} \\
\midrule
\textbf{\textit{AutoRegressive}} \\
\midrule
SEED-X~\citep{seed-x} & 0.49 & 0.86 & 0.70 &- \\
Show-o~\citep{show-o} & 0.53 & 0.70 & 0.60 &- \\
VILA-U~\citep{vila-u} &- & 0.76 & 0.64 &- \\
Emu3~\citep{emu3} & 0.54/0.65* & 0.78 & 0.60 & 80.60 \\
SimpleAR~\citep{wang2025simplear} & 0.63 &- &- & 81.97 \\
Fluid~\citep{fan2024fluid} & 0.69 &- &- &- \\
Infinity~\citep{han2025infinity} & 0.79 &- &- & 86.60 \\
Janus-Pro-7B~\citep{januspro2025} & 0.80 & 0.86 & 0.66 & 84.19 \\
Token-Shuffle~\citep{ma2025token-shuffle} & 0.62 & 0.78 & 0.67 &- \\
\rowcolor{myblue}
\Ours & 0.63/0.73$^\dag$ & 0.88/\emph{0.90}$^\dag$ & 0.67/\emph{0.74}$^\dag$ & 85.28 \\
\bottomrule
\end{tabular}}
\label{tab:bmks}
\end{table}

\begin{table}[t!]
\centering
\caption{\small Comparison on OneIG-Bench~\citep{chang2025oneigbench} in English prompts.
}
\scriptsize
\resizebox{\columnwidth}{!}{
\begin{tabular}{lcccccc}
\toprule
\textbf{Method} & \textbf{Alignment} & \textbf{Text} & \textbf{Reasoning} & \textbf{Style} & \textbf{Diversity} & \textbf{Overall$\uparrow$} \\
\midrule
\textbf{\textit{Proprietary}} \\
\midrule
Imagen3~\citep{baldridge2024imagen3} & 0.843 & 0.343 & 0.313 & 0.359 & 0.188 & 0.409 \\
Recraft V3~\citep{2024recraftv3} & 0.810 & 0.795 & 0.323 & 0.378 & 0.205 & 0.502 \\
Kolors 2.0~\citep{2025Kolors2} & 0.820 & 0.427 & 0.262 & 0.360 & 0.300 & 0.434 \\
Seedream 3.0~\citep{gao2025seedream_v3} & 0.818 & \emph{0.865} & 0.275 & 0.413 & 0.277 & 0.530 \\
Imagen4~\citep{2025Imagen4} & \emph{0.857} & 0.805 & \emph{0.338} & 0.377 & 0.199 & 0.515 \\
GPT-4o~\citep{openai2024gpt4o_image} & 0.851 & 0.857 & \textbf{0.345} & \textbf{0.462} & 0.151 & \emph{0.533} \\
\midrule
\textbf{\textit{Diffusion}} \\
\midrule
Stable Diffusion 1.5~\citep{sdv15} & 0.565 & 0.010 & 0.207 & 0.383 & \textbf{0.429} & 0.319 \\
Stable Diffusion XL~\citep{sdxl} & 0.688 & 0.029 & 0.237 & 0.332 & 0.296 & 0.316 \\
Stable Diffusion 3.5 Large~\citep{2024sd3.5} & 0.809 & 0.629 & 0.294 & 0.353 & 0.225 & 0.462 \\
Flux.1-dev~\citep{flux2024} & 0.786 & 0.523 & 0.253 & 0.368 & 0.238 & 0.434 \\
CogView4~\citep{2025cogview4} & 0.786 & 0.641 & 0.246 & 0.353 & 0.205 & 0.446 \\
SANA-1.5 1.6B (PAG)~\citep{xie2025sana} & 0.762 & 0.054 & 0.209 & 0.387 & 0.222 & 0.327 \\
SANA-1.5 4.8B (PAG)~\citep{xie2025sana} & 0.765 & 0.069 & 0.217 & 0.401 & 0.216 & 0.334 \\
Lumina-Image 2.0~\citep{qin2025lumina2} & 0.819 & 0.106 & 0.270 & 0.354 & 0.216 & 0.353 \\
HiDream-I1-Full~\citep{cai2025hidream} & 0.829 & 0.707 & 0.317 & 0.347 & 0.186 & 0.477 \\
BLIP3-o~\citep{chen2025blip3} & 0.711 & 0.013 & 0.223 & 0.361 & 0.229 & 0.307 \\
BAGEL~\citep{deng2025emerging} & 0.769 & 0.244 & 0.173 & 0.367 & 0.251 & 0.361 \\
% BAGEL $w/$ Self-CoT~\citep{deng2025emerging} & 0.793 & 0.020  & 0.206 & 0.390 & 0.209 & 0.324 \\
Show-o2-1.5B~\citep{xie2025showo2} & 0.798 & 0.002 & 0.219& 0.317 & 0.186 & 0.304 \\
Show-o2-7B~\citep{xie2025showo2} & 0.817 & 0.002 & 0.226 & 0.317 & 0.177 & 0.308 \\
OmniGen2~\citep{wu2025omnigen2} & 0.804 & 0.680 & 0.271 & 0.377 & 0.242 & 0.475 \\
Qwen-Image~\citep{wu2025qwen_image} & \textbf{0.882} & \textbf{0.891} & 0.306 & \emph{0.418} & 0.197 & \textbf{0.539} \\
\midrule
\textbf{\textit{AutoRegressive}} \\
\midrule
Emu3~\citep{emu3} & 0.737 & 0.010 & 0.193 & 0.361 & 0.251 & 0.311 \\
Janus-Pro~\citep{januspro2025} & 0.553 & 0.001 & 0.139 & 0.276 & \emph{0.365} & 0.267 \\
\rowcolor{myblue}
\Ours & 0.826 & 0.507 & 0.224 & 0.332 & 0.199 & 0.417 \\
\bottomrule
\end{tabular}
}
\label{tab:oneig_overall}
\end{table}

\section{Model Performance}

\subsection{Performance of Text-to-Image Generation}
We comprehensively evaluate the text-to-image (T2I) generation performance of \Ours on several representative benchmarks, each targeting different aspects of image generation, including visual-textual alignment and world knowledge.

\paragraph{Image–Text Alignment.}
As shown in Table~\ref{tab:bmks}, we assess \Ours's prompt-following ability across three key benchmarks. On GenEval~\citep{ghosh2023geneval}, \Ours scores \textbf{0.63} (\textbf{0.73} with Self-CoT), demonstrating robust counting, grounding, and spatial alignment. Its strong compositional abilities are further validated on GenAI-Bench~\citep{li2024evaluating}, where it achieves \textbf{0.88} on basic prompts and \textbf{0.67} on advanced prompts (\textbf{0.9} and \textbf{0.74} with Self-CoT). These results demonstrate \Ours as a great autoregressive image generation model, with performance competitive with some diffusion models such as Stable Diffusion 3.5 Large~\citep{2024sd3.5} and BAGEL~\citep{deng2025emerging}. Finally, when evaluated on DPG-Bench~\citep{hu2024ella} for long-context, multi-object scenes, \Ours achieves \textbf{85.28}, confirming its reliable compositional fidelity under complex prompts.

To perform a fine-grained analysis, we evaluated our model on OneIG-Bench~\citep{chang2025oneigbench} with English prompts. This benchmark assesses performance across areas such as alignment, text rendering, reasoning and stylistic control. As shown in \Cref{tab:oneig_overall}, \Ours achieves an overall score of \textbf{0.417}. This result significantly outperforms its autoregressive peers, such as Emu3~\citep{emu3} (\textbf{0.311}) and Janus-Pro~\citep{januspro2025} (\textbf{0.267}).

\paragraph{World Knowledge.} To evaluate \Ours's ability to integrate world knowledge into image generation, we use the WISE benchmark~\citep{niu2025wise}, which emphasizes factual grounding and semantic understanding. As shown in Table~\ref{tab:wisescore}, \Ours achieves the best performance among autoregressive models with an overall score of \textbf{0.54} (\textbf{0.67} with Self-CoT), also exceeding most diffusion models. Notably, under the prompt rewrite protocol, its score increases to \textbf{0.79} (\textbf{0.83} with Self-CoT). Collectively, these results demonstrate \Ours's robust knowledge-aware semantic alignment and cross-domain reasoning capabilities.

\begin{table}[t!]
\centering
\caption{\small Comparison of world knowledge reasoning on WISE~\citep{niu2025wise}. $\dag$ result is with Self-CoT.
% WISE examines the complex semantic understanding and world knowledge for T2I generation.
}
% \scriptsize
% \vspace{0.2cm}
\resizebox{\columnwidth}{!}{
% \begin{tabular}{lcccccccc}
\begin{tabular}{l@{ }c@{ }c@{ }c@{ }c@{ }c@{ }c@{ }c@{ }c}
\toprule
\textbf{Model} & \textbf{Cultural} & \textbf{Time} & \textbf{Space} & \textbf{Biology} & \textbf{Physics} & \textbf{Chemistry} & \textbf{Overall$\uparrow$} & \textbf{Overall (Rewrite)$\uparrow$} \\
\midrule
\textit{\textbf{Proprietary}} \\
\midrule
GPT-4o~\citep{openai2024gpt4o_image} & \textbf{0.81} & \textbf{0.71} & \textbf{0.89} & \textbf{0.83} & \textbf{0.79} & \textbf{0.74} & \textbf{0.80} &-\\
\midrule
\textit{\textbf{Diffusion}} \\
\midrule
Stable Diffusion 1.5~\citep{sdv15} & 0.34 & 0.35 & 0.32 & 0.28 & 0.29 & 0.21 & 0.32 & 0.50 \\
Stable Diffusion XL~\citep{sdxl} & 0.43  & 0.48 & 0.47 & 0.44 & 0.45 & 0.27 & 0.43 & 0.65 \\
Stable Diffusion 3.5 Large~\citep{2024sd3.5} & 0.44 & 0.50 & 0.58 & 0.44 & 0.52 & 0.31 & 0.46 & 0.72 \\
PixArt-Alpha~\citep{chen2024pixart} & 0.45 & 0.50 & 0.48 & 0.49 & 0.56 & 0.34 & 0.47 & 0.63 \\
Playground v2.5~\citep{li2024playground} & 0.49 & 0.58 & 0.55 & 0.43 & 0.48 & 0.33 & 0.49 & 0.71 \\
Flux.1-dev~\citep{flux2024} & 0.48 & 0.58 & 0.62 & 0.42 & 0.51 & 0.35 & 0.50 & 0.73 \\
MetaQuery-XL~\citep{pan2025transfer} & 0.56 & 0.55 & 0.62 & 0.49 & 0.63 & 0.41 & 0.55 &- \\
BAGEL~\citep{deng2025emerging} & 0.44/\emph{0.76}$^\dag$ & 0.55/\emph{0.69}$^\dag$ & 0.68/0.75$^\dag$ & 0.44/\emph{0.65}$^\dag$ & 0.60/\emph{0.75}$^\dag$ & 0.39/\emph{0.58}$^\dag$ & 0.52/\emph{0.70}$^\dag$ & 0.71/0.77$^\dag$ \\
% BAGEL $w/$ Self-CoT~\citep{deng2025emerging} & 0.76 & 0.69 & 0.75 & 0.65 & 0.75 & 0.58 & \emph{0.70} & 0.77 \\
Qwen-Image~\citep{wu2025qwen_image} & 0.62 & 0.63 & \emph{0.77} & 0.57 & \emph{0.75} & 0.40 & 0.62 &- \\
\midrule
\textit{\textbf{AutoRegressive}} \\
\midrule
Show-o-512~\citep{show-o} & 0.28 & 0.40 & 0.48 & 0.30 & 0.46 & 0.30 & 0.35 & 0.64 \\
VILA-U~\citep{vila-u} & 0.26 & 0.33 & 0.37 & 0.35 & 0.39 & 0.23 & 0.31 &-\\
Emu3~\citep{emu3} & 0.34 & 0.45 & 0.48 & 0.41 & 0.45 & 0.27 & 0.39 & 0.63 \\
Janus-Pro-7B~\citep{januspro2025} & 0.30 & 0.37 & 0.49 & 0.36 & 0.42 & 0.26 & 0.35 & 0.71 \\
\rowcolor{myblue} \textbf{\Ours} & 0.51/0.70$^\dag$~ & 0.54/0.65$^\dag$~ & 0.61/0.69$^\dag$~ & 0.52/0.63$^\dag$~ & 0.63/0.73$^\dag$~ & 0.48/0.52$^\dag$~  & 0.54/0.67$^\dag$~ & \emph{0.79}/\textbf{0.83}$^\dag$~ \\
\bottomrule
\end{tabular}}

\label{tab:wisescore}
\end{table}
\begin{table}[t!]
\centering
\caption{\small Comparison of image editing performance on GEdit-Bench (Full Set)~\citep{liu2025step1x} and ImgEdit-Bench~\citep{ye2025imgedit}. G\_SC, G\_PQ, and G\_O refer to the metrics evaluated by GPT-4.1~\citep{gpt4.1}. Performance is evaluated based on the \Oursedit with 1:1 aspect ratio.}
\scriptsize
% \vspace{0.2cm}
\resizebox{\columnwidth}{!}{
\begin{tabular}{lccccccc}
\toprule
\multirow{2}{*}{\textbf{Model}} & \multicolumn{3}{c}{\textbf{GEdit-Bench-EN (Full Set)$\uparrow$}} & \multicolumn{3}{c}{\textbf{GEdit-Bench-CN (Full Set)$\uparrow$}} & \multirow{2}{*}{\textbf{ImgEdit-Bench$\uparrow$}}\\
\cmidrule(lr){2-4}
\cmidrule(lr){5-7}
& \textbf{G\_SC} & \textbf{G\_PQ} & \textbf{G\_O} & \textbf{G\_SC} & \textbf{G\_PQ} & \textbf{G\_O} \\
\midrule
\textbf{\textit{Proprietary}} \\
\midrule
Gemini 2.0~\citep{gemini_2p0_flash_image_gen} & 6.87 & 7.44 & 6.51 & 5.26 & 7.60 & 5.14 &- \\
Doubao~\citep{shi2024seededit} & 7.22 & 7.89 & 6.98 & 7.17 & \emph{7.79} & 6.84 &- \\
GPT-4o~\citep{openai2024gpt4o_image} & \textbf{7.74} & \textbf{8.13} & \textbf{7.49} & \emph{7.52} & \textbf{8.02} & \textbf{7.30} & \textbf{4.20} \\
Flux.1-Kontext-pro~\citep{labs2025flux1kontextflowmatching} & 7.02 & \emph{7.60} & 6.56 & 1.11 & 7.36 & 1.23 &- \\
\midrule
\textbf{\textit{Open-source}} \\
\midrule
Instruct-Pix2Pix~\citep{Brooks2022InstructPix2PixLT} & 3.30 & 6.19 & 3.22 &- &- &- & 1.88 \\
MagicBrush~\citep{zhang2023magicbrush} & 4.52 & 6.37 & 4.19 &- &- &- & 1.83 \\
AnyEdit~\citep{yu2024anyedit} & 3.05 & 5.88 & 2.85 &- &- &- & 2.45 \\
OmniGen~\citep{xiao2024omnigen} & 5.88 & 5.87 & 5.01 &- &- &- & 2.96 \\
OmniGen2~\citep{wu2025omnigen2} & 7.16 & 6.77 & 6.41 &- &- &- & 3.44 \\
Step1X-Edit v1.0~\citep{liu2025step1x} & 7.13 & 7.00 & 6.44 & 7.30 & 7.14 & 6.66 & 3.06 \\
Step1X-Edit v1.1~\citep{liu2025step1x} & \emph{7.66} & 7.35 & 6.97 & \textbf{7.65} & 7.40 & \emph{6.98} &-\\
BAGEL~\citep{deng2025emerging} & 7.36 & 6.83 & 6.52 & 7.34 & 6.85 & 6.50 & 3.42 \\
Flux.1-Kontext-dev~\citep{labs2025flux1kontextflowmatching} &- &- & 6.26 &- &- &- & 3.71 \\
GPT-Image-Edit~\citep{wang2025gpt-edit-1_5M} &- &- & \emph{7.24} &- &- &- & \emph{3.80} \\
\rowcolor{myblue}
\Ours & 7.15 & 7.01 & 6.58 & 6.88 & 7.02 & 6.40 & 3.71 \\
\bottomrule
\end{tabular}
}

\label{tab:gedit}
% \vspace{-1pt}
\end{table}

\subsection{Performance of Image Editing}

\paragraph{Quantitative Results on Editing Benchmarks.} We developed \Oursedit by finetuning \Ours on 1M high-quality edit-only data in \Cref{sec:image-to-image-data}, demonstrates competitive performance against advanced diffusion-based models. As shown in \Cref{tab:gedit}, \Oursedit achieves scores of \textbf{6.58} on GEdit-Bench-EN~\citep{liu2025step1x} and \textbf{3.71} on ImgEdit-Bench~\citep{ye2025imgedit}, indicating its strong practical editing capabilities.

\section{Discussions}

\subsection{What Governs Image Generation: the AR Transformer or the FM Head? }
\label{sec:patch_leval_ar_diffusion}

\begin{figure}[htbp]
    \centering
    \begin{minipage}[t]{0.38\textwidth}\vspace{0pt}
        \centering
        \includegraphics[width=\linewidth]{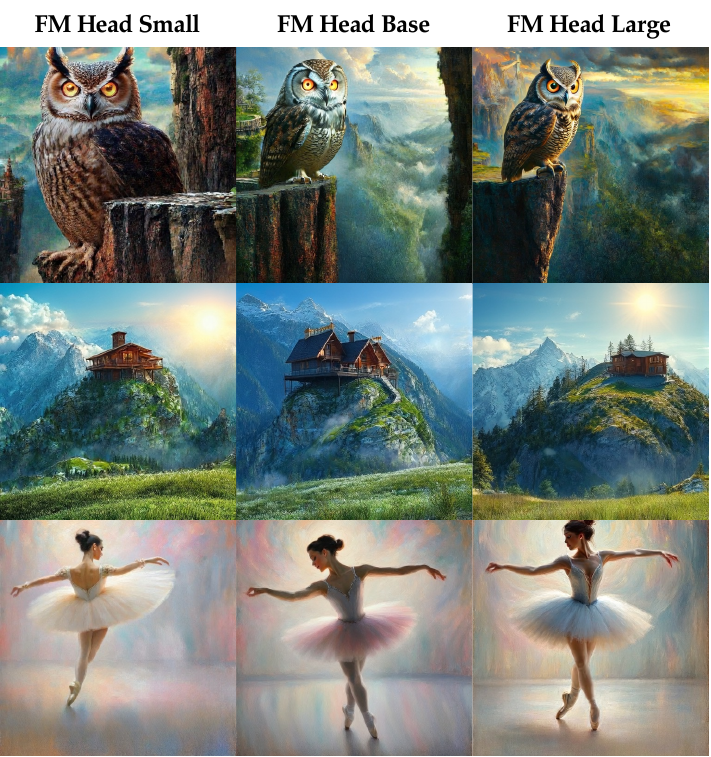}
        \caption{\small Images generated under different flow-matching heads.}
        \label{fig:head_size}
    \end{minipage}%
    \hfill
    \begin{minipage}[t]{0.58\textwidth}
        \vspace{5pt}
        \centering

        \scriptsize
        \captionof{table}{\small Configurations for different flow-matching heads.}
        % \vspace{-0.7em}
        \resizebox{\columnwidth}{!}{
        \begin{tabular}{cccc}
            \toprule
            & \textbf{Layers} & \textbf{Hidden Size} & \textbf{\# Parameters} \\
            \midrule
            FM Head Small & 6 & 1024 & 40M \\
            FM Head Base & 12 & 1536 & 157M \\
            FM Head Large & 24 & 2048 & 528M \\
            \bottomrule
        \end{tabular}}
        \label{tab:head_size}

        \scriptsize
        \captionof{table}{\small Quantitative results for different flow-matching head configurations. All variants are finetuned from the baseline with a newly initialized head.}
        \vspace{1em}
        \resizebox{\columnwidth}{!}{
        \begin{tabular}{cccc}
            \toprule
            & \textbf{GenEval} & \textbf{GenAI-Bench} & \textbf{DPG-Bench} \\
            \midrule
            Baseline & 0.59 & 0.77 & 85.15 \\
            \midrule
            w/ FM Head Small & 0.55 & 0.76 & 83.46 \\
            w/ FM Head Base & 0.55 & 0.75 & 84.68 \\
            w/ FM Head Large & 0.56 & 0.77 & 85.50 \\
            \bottomrule
        \end{tabular}}
        \label{tab:head_size_performance}
    \end{minipage}
\end{figure}

A key architectural distinction of our framework is its direct, autoregressive modeling of continuous image tokens using a flow matching objective. Prevailing autoregressive models for image generation ~\citep{emu1,emu2,dreamllm,transfusion,chen2025blip3} typically rely on heavy diffusion models for an entire image: an autoregressive model first produces a semantic embedding, which is then used to condition a diffusion model that generates an entire image in a single denoising process. In contrast, our model autoregressively generates the image patch-by-patch~\citep{kou2024orthus,chen2024diffusion,fan2025unified}, modeling the distribution of each patch with a significantly more lightweight flow matching model. We argue that this establishes our framework under the pure autoregressive paradigm with next-token prediction (NTP) modeling, rather than a diffusion model merely orchestrated by a Transformer.

A key finding from our experiments is the model's surprising insensitivity to the size of its flow-matching head. We ablated this on three heads with different sizes (small, base, and large). For each experiment, we re-initialized and trained only the head for 10k steps. Despite the significant variation in model size, all three heads produced remarkably similar results (\Cref{tab:head_size_performance}, \Cref{fig:head_size}). This insensitivity to the head's size strongly suggests that the transformer backbone performs the core generative modeling of the conditional distribution $p(x_i \mid x_{<i})$. The flow-matching head, akin to the LM head in language models, primarily acts as a lightweight sampler that translates the transformer's contextual prediction into a continuous token. Consequently, the essential generative logic resides within the transformer's autoregressive NTP process.
 
\begin{figure*}[t]
\centering
\includegraphics[width=\linewidth]{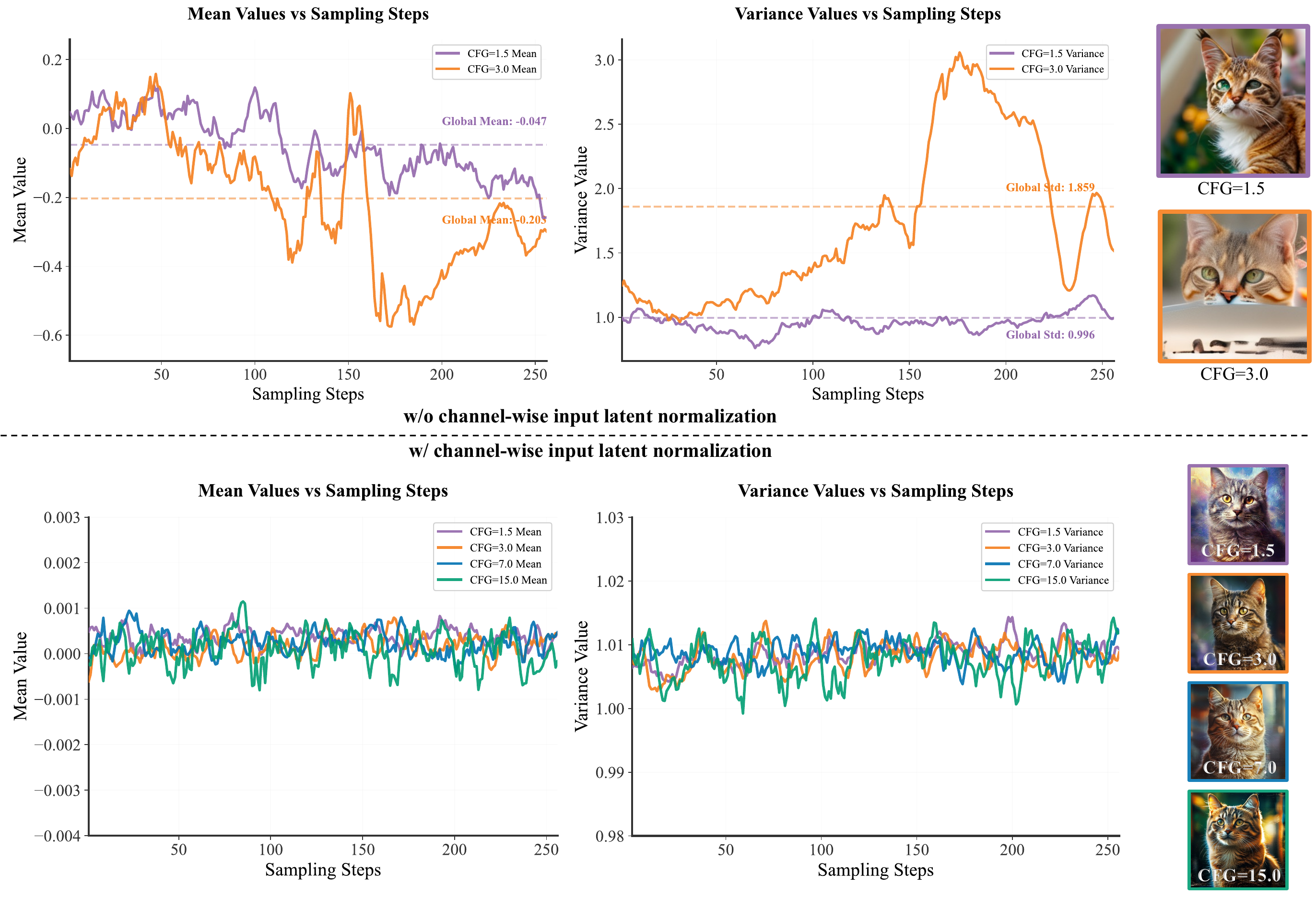}
\caption{
% Evolution of the mean and variance of output latents as the sampling step increases under two CFG settings. When CFG = 1.5, the latent distributions remain stable and close to $\mathcal{N}(0, 1)$. In contrast, when CFG = 3.0, they drift rapidly away from $\mathcal{N}(0, 1)$, leading to significant deterioration in the generated images.
\small Evolution of per-token mean and variance over sampling steps under two CFG settings. At CFG = 1.5, the mean and variance stay close to 0 and 1, respectively, indicating stability. At CFG = 3.0, they drift significantly, causing image quality degradation. With normalization, the distributions of output latents remain stable across all CFG settings.
}
\label{fig:mean_std_steps}
\end{figure*}

\subsection{Tokenizer is the Key to Image Generation}

\paragraph{Mitigating Instability under Strong Classifier-Free Guidance.}

A known failure mode in VAE-based autoregressive models is the emergence of visual artifacts, such as gray patches, particularly under strong classifier-free guidance scales~\citep{fan2024fluid}. While prior work hypothesized this instability stemmed from discontinuities in 1D positional embeddings, our analysis reveals that the root cause lies in the amplification of token-level distributional shifts under high guidance scales.

At inference time, CFG is introduced to enhance conditional fidelity. The guided prediction $\tilde{v}$ is computed via an interpolation:
\begin{equation}
\tilde{v}(x|y) = (1-w)\ \cdot v_\theta(x | \varnothing) + w \cdot v_\theta(x|y)
\end{equation} 
where $v_\theta(x | \varnothing)$ and $v_\theta(x|y)$ are the unconditional and conditional predictions, and $w$ is guidance scale. In diffusion models, inference with a high guidance scale is stable because latent variables are typically normalized, ensuring that conditional and unconditional predictions maintain a \textbf{consistent scale}. However, in token-level autoregressive models, global normalization of the entire latent tensor does not enforce per-token statistical consistency. Consequently, small discrepancies between conditional and unconditional predictions are magnified by a large guidance scale, leading to a significant drift in the statistics of generated tokens over the sequence.

We empirically demonstrate this phenomenon in \Cref{fig:mean_std_steps}. At a moderate guidance scale of 1.5, the per-token mean and variance remain stable throughout the generation process. In contrast, at a high guidance scale of 3.0, both statistics diverge significantly for later tokens, a distributional shift that corresponds directly to the appearance of visual artifacts. Our tokenizer design, which incorporates channel-wise normalization (see \Cref{eq:add_noise}), directly addresses this issue by enforcing per-token statistical stability. This simple but critical design choice mitigates the instability, enabling the use of strong guidance without degrading image quality.

\paragraph{A Regularized Latent Space is Critical for Generation}

\begin{figure*}[t!]
\centering
\includegraphics[width=\linewidth]{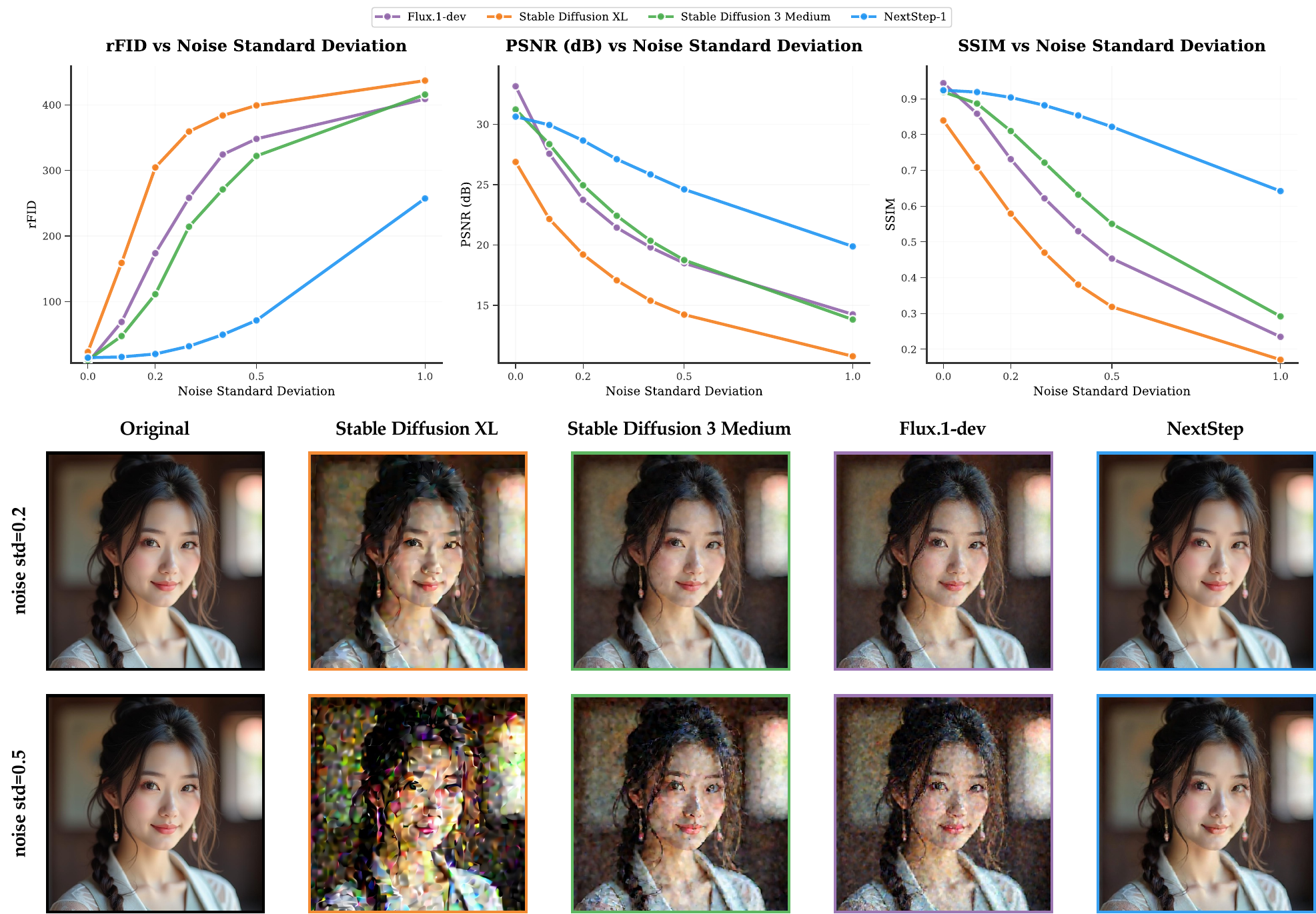}
% \caption{\small Impact of Noise Injection on Image Tokenizer Performance. The top panel displays the quantitative evaluation of different image tokenizers, plotting rFID$\downarrow$, PSNR$\uparrow$, and SSIM$\uparrow$ as a function of injected noise intensity. The bottom panel presents qualitative comparisons of image reconstruction quality for the same image tokenizer, specifically at noise standard deviations of 0.2 and 0.5.}
\caption{\small Impact of Noise Perturbation on Image Tokenizer Performance. The top panel displays quantitative metrics (rFID$\downarrow$, PSNR$\uparrow$, and SSIM$\uparrow$) versus noise intensity. The bottom panel presents qualitative reconstruction examples at noise standard deviations of 0.2 and 0.5.}
\label{fig:vae_robustness}
\end{figure*}

\begin{figure*}[t!]
\centering
\includegraphics[width=1\linewidth]{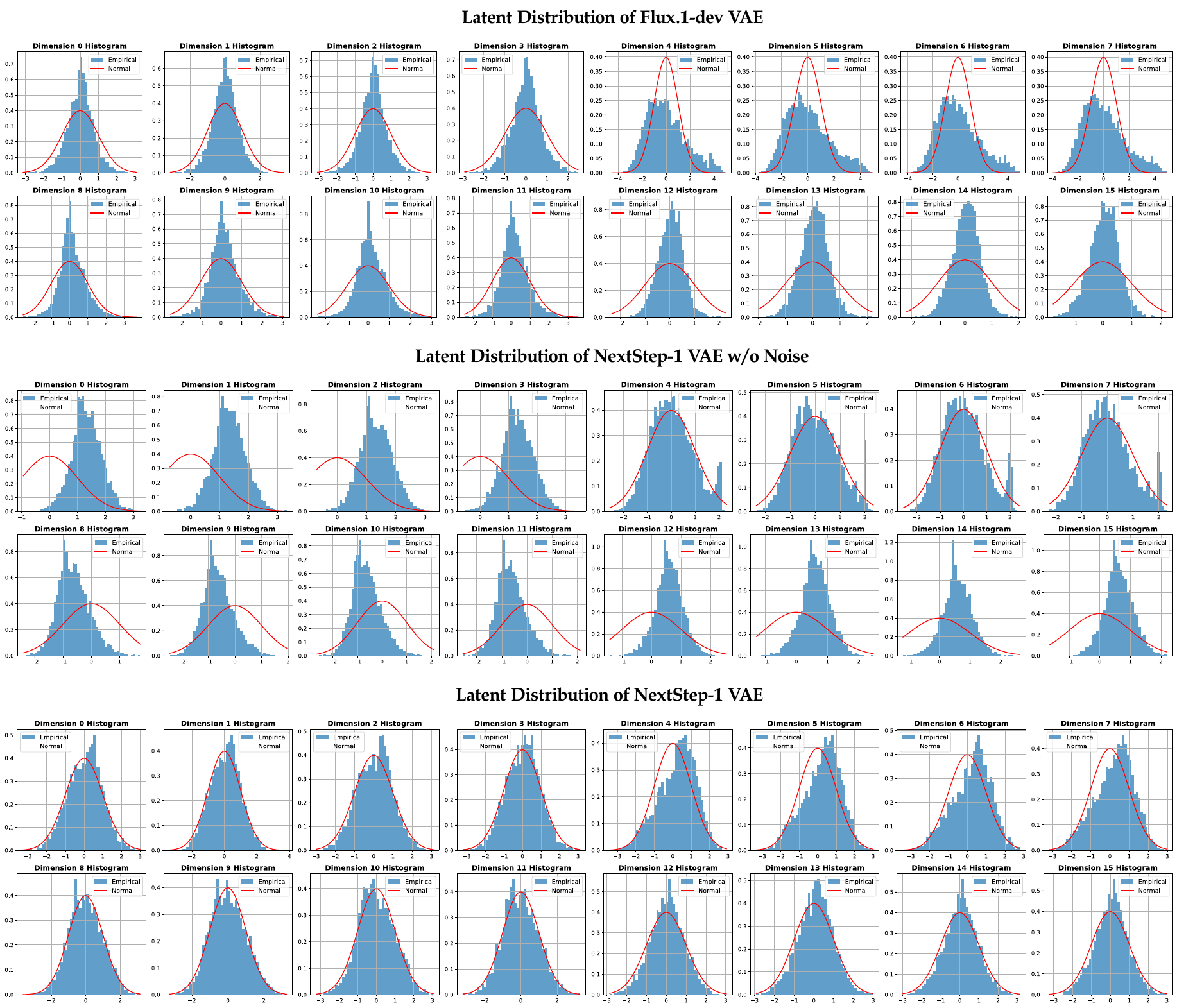}
\caption{
\small Latent distributions in 16 channels for three VAE variants: Flux.1-dev, NextStep-1 w/o noise, and NextStep-1. Blue bars show empirical histograms; red lines indicate the standard normal distribution. NextStep-1 VAE aligns best with the normal distribution, reflecting a dispersed latent distribution.}
\label{fig:dist}
\end{figure*}

A key finding of our work is a counterintuitive inverse correlation between the generation loss and the final synthesis quality of the autoregressive model. Specifically, applying higher noise intensity ($\gamma$ in \Cref{eq:add_noise}) during tokenizer training increases generation loss but paradoxically improves the quality of the generated images. For instance, \Ours uses a tokenizer trained at $\gamma=0.5$, which incurred the highest generation loss yet produced the highest-fidelity images. 
Conversely, tokenizers trained for low generation loss caused the autoregressive model to yield outputs resembling pure noise.

We attribute this phenomenon to noise regularization, cultivating a well-conditioned latent space. This process enhances two key properties: the tokenizer decoder's robustness to latent perturbations (\Cref{fig:vae_robustness}) and a more dispersed latent distribution (\Cref{fig:dist}), a property prior work has also found beneficial for generation~\citep{yang2025latent_make_good_tokenizer,yao2025vavae,sun2024latentlm}. While it remains unclear whether robustness or dispersion plays a critical role, these results underscore the practical benefits of noise-based regularization and highlight promising directions for future analysis.

\paragraph{Reconstruction Quality is the Upper Bound of Generation Quality.}

\begin{table}[t!]
\centering
\scriptsize
\caption{\small Comparison of reconstruction performance on ImageNet-1K 256$\times$256~\citep{deng2009imagenet}.}
% \vspace{0.2cm}
\begin{tabular}{lccc}
\toprule
\textbf{Tokenizer} & \textbf{Latent Shape} & \textbf{PSNR~$\uparrow$} & \textbf{SSIM~$\uparrow$} \\
\midrule
\multicolumn{4}{l}{\textit{\textbf{Discrete Tokenizer}}} \\
\midrule
SBER-MoVQGAN (270M)~\citep{movqgan} & 32x32 & 27.04 & 0.74 \\
LlamaGen~\citep{sun2024llamagen} & 32x32 & 24.44 & 0.77 \\
VAR~\citep{tian2024visual} & 680 & 22.12 & 0.62 \\
TiTok-S-128~\citep{yu2024titok} & 128 & 17.52 & 0.44 \\
Sefltok~\citep{wang2025selftok} & 1024 & 26.30 & 0.81 \\
\midrule
\multicolumn{4}{l}{\textit{\textbf{Continuous Tokenizer}}} \\
\midrule
Stable Diffusion 1.5~\citep{sdv15} & 32x32x4 & 25.18 & 0.73 \\
Stable Diffusion XL~\citep{sdxl} & 32x32x4 & 26.22 & 0.77 \\
Stable Diffusion 3 Medium~\citep{SD3} & 32x32x16 & 30.00 & 0.88 \\
Flux.1-dev~\citep{flux2024} & 32x32x16 & \textbf{31.64} & \textbf{0.91} \\
\rowcolor{myblue}
\textbf{\Ours} & 32x32x16 & \emph{30.60} & \emph{0.89} \\
\bottomrule
\end{tabular}
\label{tab:vae_results}
\end{table}

The reconstruction fidelity of the image tokenizer fundamentally determines the upper bound for the quality of the final generated image, particularly for fine details and textures. This principle has been validated in numerous recent studies~\citep{SD3, flux2024, dai2023meta_emu}, leading to a trend in the diffusion paradigm of building generative models on top of VAEs with exceptional reconstruction performance (e.g., PSNR > 30). In contrast, VQ-based autoregressive models have historically struggled to surpass this threshold, as shown in \Cref{tab:vae_results}. While a trade-off between reconstruction and generation quality is often debated~\citep{yao2025vavae}, our work successfully applies autoregressive models to high-fidelity continuous VAEs, bridging this gap.

% \newpage
\subsection{Limitations and Challenges}

\begin{figure*}[t!]
\centering
\includegraphics[width=\linewidth]{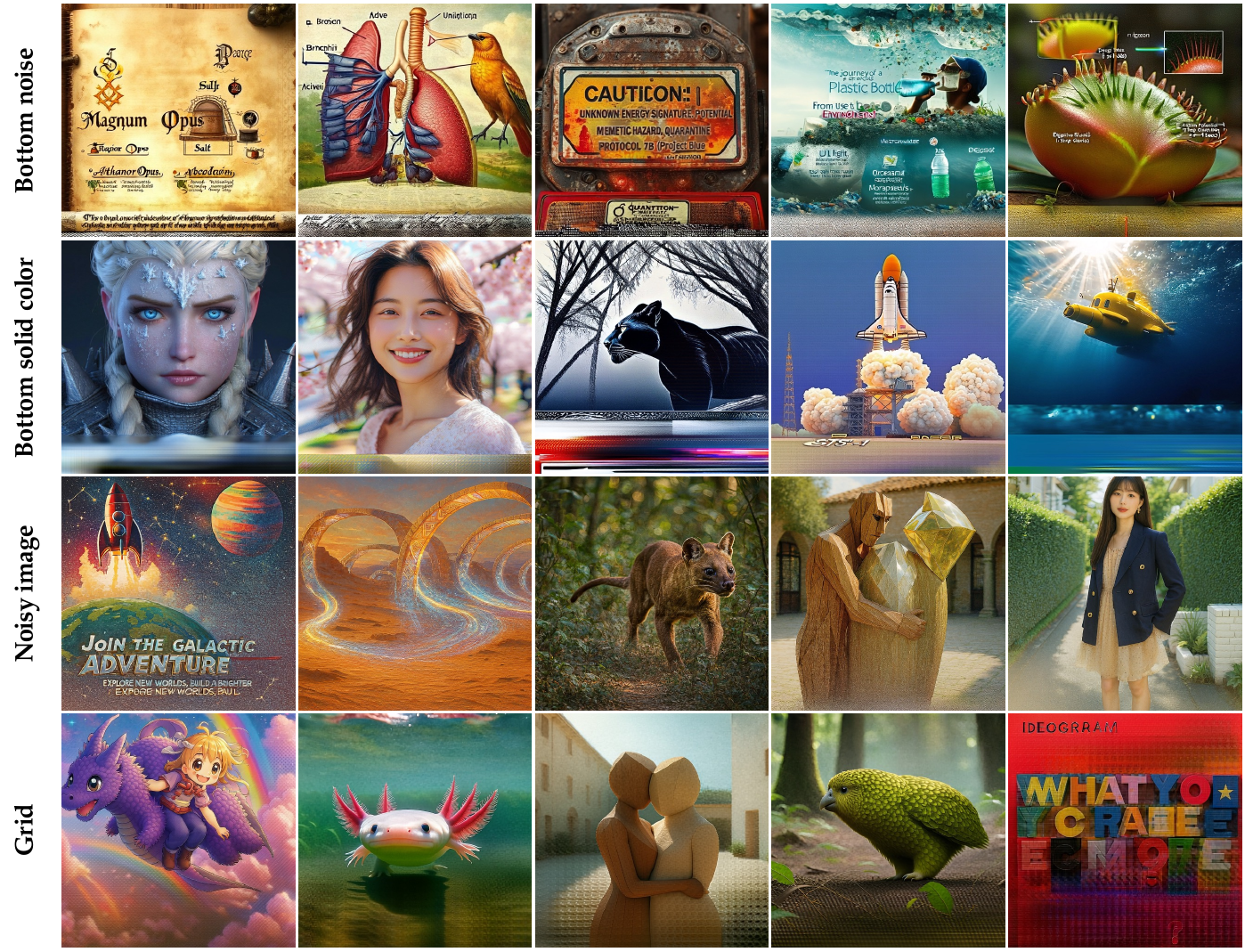}
\caption{
\small Failure cases for high-dimensional continuous tokens.
}
\label{fig:failure_case}
\end{figure*}

\paragraph{Artifacts.} While \Ours successfully demonstrates that autoregressive models can operate on high-dimensional continuous latent spaces, achieving generation quality comparable to diffusion models, this approach also introduces unique stability challenges. We observed the emergence of several distinct generative artifacts when transitioning from a VAE with a lower-dimensional latent space (e.g., spatial downsample factor is 8 and number of latent channels is 4) to one with a higher-dimensional space (e.g., spatial downsample factor is 8 and number of latent channels is 16). While the former configuration produced stable outputs, the latter occasionally exhibited failure modes, as illustrated in \Cref{fig:failure_case}.

While the underlying causes remain an open question, we identify several plausible contributing factors: (1) \textbf{Local noise or block-shaped artifacts} emerging in the later stages of generation may arise from numerical instabilities; (2) \textbf{Global noise across the image} may reflect under-convergence, implying that additional training could mitigate the issue; and (3) \textbf{Subtle grid-like artifacts} could reveal limitations of the 1D positional encoding in capturing 2D spatial relationships.

\paragraph{Inference Latency of Sequential Decoding.}
A theoretical analysis of per-token latency on an H100 GPU (983 TFLOPS, 3.36 TB/s bandwidth) with a batch size of 1, as detailed in \Cref{tab:inf}, decomposes the contributions of individual components. The results show that the dominant bottleneck lies in the serial decoding of the LLM, while the multi-step sampling in the flow-matching head also constitutes a substantial portion of the per-token generation cost. These observations suggest two promising directions for accelerating inference. First, the efficiency of the flow matching head could be improved by reducing its parameter count, applying distillation to achieve few-step generation~\citep{meng2023distillation}, or using more advanced few-step samplers~\citep{lu2022dpm_solver,lu2025dpm_plus}. Second, the autoregressive backbone could be accelerated by adapting recent advances from the LLM field, such as speculative decoding~\citep{leviathan2023speculative} or multi-token prediction~\citep{gloeckle2024mtp}, to the domain of image token generation.

\begin{table}[ht]
\centering
\scriptsize
\caption{\small Inference latency breakdown at 983 TFLOP/s compute and 3.36~TB/s memory bandwidth.}
\begin{tabular}{lccccc}
\toprule
\multirow{2}{*}{\textbf{Sequence Length}} & \multicolumn{3}{c}{\textbf{Last-token Latency (ms)}} & \multicolumn{2}{c}{\textbf{Accumulated Latency (s)}}\\
\cmidrule(lr){2-4}
\cmidrule(lr){5-6}
% \midrule
 & \textbf{LLM Decoder} & \textbf{LM Head} & \textbf{FM Head} & \textbf{Total} & \textbf{w/o FM Head} \\
\midrule
256  & 7.20 & 0.40 & 3.40 & 2.82 & 1.95 \\
1024 & 7.23 & 0.40 & 3.40 & 11.31 & 7.83 \\
4096 & 7.39 & 0.40 & 3.40 & 45.77 & 31.86 \\
\bottomrule
\end{tabular}
\label{tab:inf}
\end{table}

\paragraph{Challenges in High-Resolution Training.} Our framework faces two primary challenges in scaling to high-resolution image generation, particularly when compared to diffusion models, which benefit from well-established techniques~\citep{SD3,chen2024pixart} in this domain. First, the strictly sequential nature of autoregressive generation requires substantially more training steps to converge at higher resolutions. In contrast, diffusion models refine the entire image in parallel at each iteration, enabling more direct exploitation of 2D spatial inductive biases. Second, techniques recently developed for high-resolution diffusion models, such as timestep shift, are difficult to adapt to our setting. This limitation arises because the Flow Matching Head acts primarily as a lightweight sampler, while the transformer backbone performs the core generative modeling; thus, modifications to the sampling process have only a marginal impact on the final output. Designing high-resolution generation strategies specifically for patch-wise autoregressive models remains an important direction for future research.

\paragraph{Challenges in SFT.} SFT in our autoregressive framework poses unique challenges compared to diffusion models. \textbf{We observe that fine-tuning on small, high-quality datasets exhibits unstable dynamics.} In contrast to diffusion models, which can often adapt to a target distribution and maintain stable and general image generation with only a few thousand samples, our SFT process yields substantial improvements only when trained on datasets at the million-sample scale. With smaller datasets, the model remains in a precarious equilibrium; it either improves marginally with negligible impact or abruptly overfits to the target distribution. Consequently, identifying an intermediate checkpoint that achieves alignment with the target distribution while preserving general generative capability remains a significant challenge.
\newpage
\section*{Contributors and Acknowledgments}
% \small{

% We designate Research as those who have participated in the development of \Ours throughout its entire process, while contributors are those who worked on the early versions or contributed part-time. $^\star$~Denotes co-first authors, and $\dagger$~denotes project leaders. Authors are listed alphabetically by first name.

We designate researchers as those who are involved in the development of \Ours, while contributors refer to those who provide support in areas such as data, systems, platforms, early version work, or part-time contributions. $^\star$ indicates core executors, and $\dagger$ indicates the project leader. Authors are listed alphabetically by first name.

\paragraph{Researchers:}
Chunrui Han$^\star$, Guopeng Li$^\star$, Jingwei Wu$^\star$, Quan Sun$^\star\dagger$, Yan Cai$^\star$, Yuang Peng$^\star$, Zheng Ge$^\star\dagger$, Deyu Zhou, Haomiao Tang, Hongyu Zhou, Kenkun Liu

\paragraph{Contributors:}
Ailin Huang, Bin Wang, Changxin Miao, Deshan Sun, En Yu, Fukun Yin, Gang Yu, Hao Nie, Haoran Lv, Hanpeng Hu, Jia Wang, Jian Zhou, Jianjian Sun, Kaijun Tan, Kang An, Kangheng Lin, Liang Zhao, Mei Chen, Peng Xing, Rui Wang, Shiyu Liu, Shutao Xia, Tianhao You, Wei Ji, Xianfang Zeng, Xin Han, Xuelin Zhang, Yana Wei, Yanming Xu, Yimin Jiang, Yingming Wang, Yu Zhou, Yucheng Han, Ziyang Meng

\paragraph{Sponsors:}
Binxing Jiao, Daxin Jiang, Xiangyu Zhang, Yibo Zhu

\paragraph{Acknowledgments:}
We would like to sincerely thank Tianhong Li and Yonglong Tian for their insightful discussions.

% }

\newpage
\bibliography{main}

\begin{thebibliography}{105}
\providecommand{\natexlab}[1]{#1}
\providecommand{\url}[1]{\texttt{#1}}
\expandafter\ifx\csname urlstyle\endcsname\relax
  \providecommand{\doi}[1]{doi: #1}\else
  \providecommand{\doi}{doi: \begingroup \urlstyle{rm}\Url}\fi

\bibitem[Bai et~al.(2025)Bai, Chen, Liu, Wang, Ge, Song, Dang, Wang, Wang, Tang, Zhong, Zhu, Yang, Li, Wan, Wang, Ding, Fu, Xu, Ye, Zhang, Xie, Cheng, Zhang, Yang, Xu, and Lin]{qwen2.5-vl}
S.~Bai, K.~Chen, X.~Liu, J.~Wang, W.~Ge, S.~Song, K.~Dang, P.~Wang, S.~Wang, J.~Tang, H.~Zhong, Y.~Zhu, M.~Yang, Z.~Li, J.~Wan, P.~Wang, W.~Ding, Z.~Fu, Y.~Xu, J.~Ye, X.~Zhang, T.~Xie, Z.~Cheng, H.~Zhang, Z.~Yang, H.~Xu, and J.~Lin.
\newblock Qwen2.5-vl technical report.
\newblock \emph{arXiv preprint arXiv:2502.13923}, 2025.

\bibitem[Baldridge et~al.(2024)Baldridge, Bauer, Bhutani, Brichtova, Bunner, Castrejon, Chan, Chen, Dieleman, Du, et~al.]{baldridge2024imagen3}
J.~Baldridge, J.~Bauer, M.~Bhutani, N.~Brichtova, A.~Bunner, L.~Castrejon, K.~Chan, Y.~Chen, S.~Dieleman, Y.~Du, et~al.
\newblock Imagen 3.
\newblock \emph{arXiv preprint arXiv:2408.07009}, 2024.

\bibitem[Betker et~al.(2023)Betker, Goh, Jing, Brooks, Wang, Li, Ouyang, Zhuang, Lee, Guo, et~al.]{dalle3}
J.~Betker, G.~Goh, L.~Jing, T.~Brooks, J.~Wang, L.~Li, L.~Ouyang, J.~Zhuang, J.~Lee, Y.~Guo, et~al.
\newblock Improving image generation with better captions.
\newblock \emph{OpenAI blog}, 2023.

\bibitem[Brooks et~al.(2023)Brooks, Holynski, and Efros]{Brooks2022InstructPix2PixLT}
T.~Brooks, A.~Holynski, and A.~A. Efros.
\newblock Instructpix2pix: Learning to follow image editing instructions.
\newblock In \emph{Proceedings of the IEEE/CVF Conference on Computer Vision and Pattern Recognition (CVPR)}, 2023.

\bibitem[Brown et~al.(2020)Brown, Mann, Ryder, Subbiah, Kaplan, Dhariwal, Neelakantan, Shyam, Sastry, Askell, et~al.]{gpt3}
T.~Brown, B.~Mann, N.~Ryder, M.~Subbiah, J.~D. Kaplan, P.~Dhariwal, A.~Neelakantan, P.~Shyam, G.~Sastry, A.~Askell, et~al.
\newblock Language models are few-shot learners.
\newblock \emph{Advances in neural information processing systems (NeurIPS)}, 2020.

\bibitem[Cai et~al.(2025)Cai, Chen, Chen, Li, Long, Pan, Qiu, Zhang, Gao, Xu, et~al.]{cai2025hidream}
Q.~Cai, J.~Chen, Y.~Chen, Y.~Li, F.~Long, Y.~Pan, Z.~Qiu, Y.~Zhang, F.~Gao, P.~Xu, et~al.
\newblock Hidream-i1: A high-efficient image generative foundation model with sparse diffusion transformer.
\newblock \emph{arXiv preprint arXiv:2505.22705}, 2025.

\bibitem[Chang et~al.(2025)Chang, Fang, Xing, Wu, Cheng, Wang, Zeng, Yu, and Chen]{chang2025oneigbench}
J.~Chang, Y.~Fang, P.~Xing, S.~Wu, W.~Cheng, R.~Wang, X.~Zeng, G.~Yu, and H.-B. Chen.
\newblock Oneig-bench: Omni-dimensional nuanced evaluation for image generation.
\newblock \emph{arXiv preprint arXiv:2506.07977}, 2025.

\bibitem[Chen et~al.(2024{\natexlab{a}})Chen, Mart{\'\i}~Mons{\'o}, Du, Simchowitz, Tedrake, and Sitzmann]{chen2024diffusion}
B.~Chen, D.~Mart{\'\i}~Mons{\'o}, Y.~Du, M.~Simchowitz, R.~Tedrake, and V.~Sitzmann.
\newblock Diffusion forcing: Next-token prediction meets full-sequence diffusion.
\newblock \emph{Advances in neural information processing systems (NeurIPS)}, 2024{\natexlab{a}}.

\bibitem[Chen et~al.(2024{\natexlab{b}})Chen, Ge, Xie, Wu, Yao, Ren, Wang, Luo, Lu, and Li]{chen2024pixart}
J.~Chen, C.~Ge, E.~Xie, Y.~Wu, L.~Yao, X.~Ren, Z.~Wang, P.~Luo, H.~Lu, and Z.~Li.
\newblock Pixart-sigma: Weak-to-strong training of diffusion transformer for 4k text-to-image generation.
\newblock In \emph{European Conference on Computer Vision (ECCV)}, 2024{\natexlab{b}}.

\bibitem[Chen et~al.(2025{\natexlab{a}})Chen, Xu, Pan, Hu, Qin, Goldstein, Huang, Zhou, Xie, Savarese, et~al.]{chen2025blip3}
J.~Chen, Z.~Xu, X.~Pan, Y.~Hu, C.~Qin, T.~Goldstein, L.~Huang, T.~Zhou, S.~Xie, S.~Savarese, et~al.
\newblock Blip3-o: A family of fully open unified multimodal models-architecture, training and dataset.
\newblock \emph{arXiv preprint arXiv:2505.09568}, 2025{\natexlab{a}}.

\bibitem[Chen et~al.(2025{\natexlab{b}})Chen, Wu, Wu, Ma, Liu, Pan, Liu, Xie, Yu, Ruan, and Luo]{januspro2025}
X.~Chen, C.~Wu, Z.~Wu, Y.~Ma, X.~Liu, Z.~Pan, W.~Liu, Z.~Xie, X.~Yu, C.~Ruan, and P.~Luo.
\newblock Janus-pro: Unified multimodal understanding and generation with data and model scaling.
\newblock \emph{arXiv preprint arXiv:2501.17811}, 2025{\natexlab{b}}.

\bibitem[Dai et~al.(2023)Dai, Hou, Ma, Tsai, Wang, Wang, Zhang, Vandenhende, Wang, Dubey, et~al.]{dai2023meta_emu}
X.~Dai, J.~Hou, C.-Y. Ma, S.~Tsai, J.~Wang, R.~Wang, P.~Zhang, S.~Vandenhende, X.~Wang, A.~Dubey, et~al.
\newblock Emu: Enhancing image generation models using photogenic needles in a haystack.
\newblock \emph{arXiv preprint arXiv:2309.15807}, 2023.

\bibitem[deepmind Imagen4~team(2025)]{2025Imagen4}
G.~deepmind Imagen4~team.
\newblock Imagen4, 2025.
\newblock URL \url{https://storage.googleapis.com/deepmind-media/Model-Cards/Imagen-4-Model-Card.pdf}.

\bibitem[Deitke et~al.(2023)Deitke, Liu, Wallingford, Ngo, Michel, Kusupati, Fan, Laforte, Voleti, Gadre, et~al.]{deitke2023objaverse}
M.~Deitke, R.~Liu, M.~Wallingford, H.~Ngo, O.~Michel, A.~Kusupati, A.~Fan, C.~Laforte, V.~Voleti, S.~Y. Gadre, et~al.
\newblock Objaverse-xl: A universe of 10m+ 3d objects.
\newblock \emph{Advances in Neural Information Processing Systems (NeurIPS)}, 2023.

\bibitem[Deng et~al.(2025)Deng, Zhu, Li, Gou, Li, Wang, Zhong, Yu, Nie, Song, et~al.]{deng2025emerging}
C.~Deng, D.~Zhu, K.~Li, C.~Gou, F.~Li, Z.~Wang, S.~Zhong, W.~Yu, X.~Nie, Z.~Song, et~al.
\newblock Emerging properties in unified multimodal pretraining.
\newblock \emph{arXiv preprint arXiv:2505.14683}, 2025.

\bibitem[Deng et~al.(2009)Deng, Dong, Socher, Li, Li, and Fei-Fei]{deng2009imagenet}
J.~Deng, W.~Dong, R.~Socher, L.-J. Li, K.~Li, and L.~Fei-Fei.
\newblock Imagenet: A large-scale hierarchical image database.
\newblock In \emph{Proceedings of the IEEE/CVF Conference on Computer Vision and Pattern Recognition (CVPR)}, 2009.

\bibitem[Dong et~al.(2024)Dong, Han, Peng, Qi, Ge, Yang, Zhao, Sun, Zhou, Wei, et~al.]{dreamllm}
R.~Dong, C.~Han, Y.~Peng, Z.~Qi, Z.~Ge, J.~Yang, L.~Zhao, J.~Sun, H.~Zhou, H.~Wei, et~al.
\newblock Dreamllm: Synergistic multimodal comprehension and creation.
\newblock In \emph{International Conference on Learning Representations (ICLR)}, 2024.

\bibitem[Eslami et~al.(2021)Eslami, Liu, Oord, Vinyals, Wainwright, and Sutskever]{eslami2021taming}
S.~M.~A. Eslami, S.~Liu, A.~v.~d. Oord, O.~Vinyals, M.~J. Wainwright, and I.~Sutskever.
\newblock Taming transformers for high-resolution image synthesis.
\newblock In \emph{International Conference on Machine Learning (ICML)}, 2021.

\bibitem[Esser et~al.(2024)Esser, Kulal, Blattmann, Entezari, M{\"u}ller, Saini, Levi, Lorenz, Sauer, Boesel, et~al.]{SD3}
P.~Esser, S.~Kulal, A.~Blattmann, R.~Entezari, J.~M{\"u}ller, H.~Saini, Y.~Levi, D.~Lorenz, A.~Sauer, F.~Boesel, et~al.
\newblock Scaling rectified flow transformers for high-resolution image synthesis.
\newblock In \emph{International Conference on Machine Learning (ICML)}, 2024.

\bibitem[Fan et~al.(2024)Fan, Li, Qin, Li, Sun, Rubinstein, Sun, He, and Tian]{fan2024fluid}
L.~Fan, T.~Li, S.~Qin, Y.~Li, C.~Sun, M.~Rubinstein, D.~Sun, K.~He, and Y.~Tian.
\newblock Fluid: Scaling autoregressive text-to-image generative models with continuous tokens.
\newblock \emph{arXiv preprint arXiv:2410.13863}, 2024.

\bibitem[Fan et~al.(2025)Fan, Tang, Qin, Li, Yang, Qiao, Steiner, Sun, Li, Zhu, et~al.]{fan2025unified}
L.~Fan, L.~Tang, S.~Qin, T.~Li, X.~Yang, S.~Qiao, A.~Steiner, C.~Sun, Y.~Li, T.~Zhu, et~al.
\newblock Unified autoregressive visual generation and understanding with continuous tokens.
\newblock \emph{arXiv preprint arXiv:2503.13436}, 2025.

\bibitem[Gao et~al.(2025)Gao, Gong, Guo, Hou, Lai, Li, Li, Lian, Liao, Liu, et~al.]{gao2025seedream_v3}
Y.~Gao, L.~Gong, Q.~Guo, X.~Hou, Z.~Lai, F.~Li, L.~Li, X.~Lian, C.~Liao, L.~Liu, et~al.
\newblock Seedream 3.0 technical report.
\newblock \emph{arXiv preprint arXiv:2504.11346}, 2025.

\bibitem[Ge et~al.(2024)Ge, Zhao, Zhu, Ge, Yi, Song, Li, Ding, and Shan]{seed-x}
Y.~Ge, S.~Zhao, J.~Zhu, Y.~Ge, K.~Yi, L.~Song, C.~Li, X.~Ding, and Y.~Shan.
\newblock Seed-x: Multimodal models with unified multi-granularity comprehension and generation.
\newblock \emph{arXiv preprint arxiv:2404.14396}, 2024.

\bibitem[Gemini2(2025)]{gemini_2p0_flash_image_gen}
G.~Gemini2.
\newblock Experiment with gemini 2.0 flash native image generation, 2025.
\newblock URL \url{https://developers.googleblog.com/en/experiment-with-gemini-20-flash-native-image-generation}.

\bibitem[Ghosh et~al.(2023)Ghosh, Hajishirzi, and Schmidt]{ghosh2023geneval}
D.~Ghosh, H.~Hajishirzi, and L.~Schmidt.
\newblock Geneval: An object-focused framework for evaluating text-to-image alignment.
\newblock In \emph{Advances in neural information processing systems (NeurIPS)}, 2023.

\bibitem[Gloeckle et~al.(2024)Gloeckle, Idrissi, Rozi{\`e}re, Lopez-Paz, and Synnaeve]{gloeckle2024mtp}
F.~Gloeckle, B.~Y. Idrissi, B.~Rozi{\`e}re, D.~Lopez-Paz, and G.~Synnaeve.
\newblock Better \& faster large language models via multi-token prediction.
\newblock \emph{arXiv preprint arXiv:2404.19737}, 2024.

\bibitem[Han et~al.(2025)Han, Liu, Jiang, Yan, Zhang, Yuan, Peng, and Liu]{han2025infinity}
J.~Han, J.~Liu, Y.~Jiang, B.~Yan, Y.~Zhang, Z.~Yuan, B.~Peng, and X.~Liu.
\newblock Infinity: Scaling bitwise autoregressive modeling for high-resolution image synthesis.
\newblock In \emph{Proceedings of the Computer Vision and Pattern Recognition Conference (CVPR)}, 2025.

\bibitem[Han et~al.(2024)Han, Wu, Shi, Liu, Liao, Qiu, Yuan, Gu, Dong, and Cui]{han2024mvimgnet2}
X.~Han, Y.~Wu, L.~Shi, H.~Liu, H.~Liao, L.~Qiu, W.~Yuan, X.~Gu, Z.~Dong, and S.~Cui.
\newblock Mvimgnet2. 0: A larger-scale dataset of multi-view images.
\newblock \emph{arXiv preprint arXiv:2412.01430}, 2024.

\bibitem[Hu et~al.(2024)Hu, Wang, Fang, Fu, Cheng, and Yu]{hu2024ella}
X.~Hu, R.~Wang, Y.~Fang, B.~Fu, P.~Cheng, and G.~Yu.
\newblock Ella: Equip diffusion models with llm for enhanced semantic alignment.
\newblock \emph{arXiv preprint arXiv:2403.05135}, 2024.

\bibitem[Kirillov et~al.(2023)Kirillov, Mintun, Ravi, Mao, Rolland, Gustafson, Xiao, Whitehead, Berg, Lo, Doll{\' a}r, and Girshick]{kirillov2023segment}
A.~Kirillov, E.~Mintun, N.~Ravi, H.~Mao, C.~Rolland, L.~Gustafson, T.~Xiao, S.~Whitehead, A.~C. Berg, W.-Y. Lo, P.~Doll{\' a}r, and R.~B. Girshick.
\newblock Segment {Anything}.
\newblock In \emph{IEEE {International} {Conference} on {Computer} {Vision} ({ICCV})}, 2023.

\bibitem[Kou et~al.(2024)Kou, Jin, Liu, Liu, Ma, Jia, Chen, Jiang, and Deng]{kou2024orthus}
S.~Kou, J.~Jin, Z.~Liu, C.~Liu, Y.~Ma, J.~Jia, Q.~Chen, P.~Jiang, and Z.~Deng.
\newblock Orthus: Autoregressive interleaved image-text generation with modality-specific heads.
\newblock \emph{arXiv preprint arXiv:2412.00127}, 2024.

\bibitem[Labs(2024)]{flux2024}
B.~F. Labs.
\newblock Flux, 2024.
\newblock URL \url{https://github.com/black-forest-labs/flux}.

\bibitem[Labs(2025)]{fluxfill}
B.~F. Labs.
\newblock Flux.1-fill-dev, 2025.
\newblock URL \url{https://huggingface.co/black-forest-labs/FLUX.1-Fill-dev}.

\bibitem[Labs et~al.(2025)Labs, Batifol, Blattmann, Boesel, Consul, Diagne, Dockhorn, English, English, Esser, et~al.]{labs2025flux1kontextflowmatching}
B.~F. Labs, S.~Batifol, A.~Blattmann, F.~Boesel, S.~Consul, C.~Diagne, T.~Dockhorn, J.~English, Z.~English, P.~Esser, et~al.
\newblock Flux. 1 kontext: Flow matching for in-context image generation and editing in latent space.
\newblock \emph{arXiv preprint arXiv:2506.15742}, 2025.

\bibitem[Leviathan et~al.(2023)Leviathan, Kalman, and Matias]{leviathan2023speculative}
Y.~Leviathan, M.~Kalman, and Y.~Matias.
\newblock Fast inference from transformers via speculative decoding.
\newblock In \emph{International Conference on Machine Learning}, pages 19274--19286. PMLR, 2023.

\bibitem[Li et~al.(2024{\natexlab{a}})Li, Lin, Pathak, Li, Fei, Wu, Xia, Zhang, Neubig, and Ramanan]{li2024evaluating}
B.~Li, Z.~Lin, D.~Pathak, J.~Li, Y.~Fei, K.~Wu, X.~Xia, P.~Zhang, G.~Neubig, and D.~Ramanan.
\newblock Evaluating and improving compositional text-to-visual generation.
\newblock In \emph{Proceedings of the IEEE/CVF Conference on Computer Vision and Pattern Recognition (CVPR)}, 2024{\natexlab{a}}.

\bibitem[Li et~al.(2024{\natexlab{b}})Li, Kamko, Akhgari, Sabet, Xu, and Doshi]{li2024playground}
D.~Li, A.~Kamko, E.~Akhgari, A.~Sabet, L.~Xu, and S.~Doshi.
\newblock Playground v2. 5: Three insights towards enhancing aesthetic quality in text-to-image generation.
\newblock \emph{arXiv preprint arXiv:2402.17245}, 2024{\natexlab{b}}.

\bibitem[Li et~al.(2024{\natexlab{c}})Li, Tian, Li, Deng, and He]{mar}
T.~Li, Y.~Tian, H.~Li, M.~Deng, and K.~He.
\newblock Autoregressive image generation without vector quantization.
\newblock In \emph{Advances in neural information processing systems (NeurIPS)}, 2024{\natexlab{c}}.

\bibitem[Liao et~al.(2025)Liao, Liu, Wang, Luo, Zhang, Zhao, Wu, Li, Tian, and Huang]{liao2025mogao}
C.~Liao, L.~Liu, X.~Wang, Z.~Luo, X.~Zhang, W.~Zhao, J.~Wu, L.~Li, Z.~Tian, and W.~Huang.
\newblock Mogao: An omni foundation model for interleaved multi-modal generation.
\newblock \emph{arXiv preprint arXiv:2505.05472}, 2025.

\bibitem[Lin et~al.(2024)Lin, Pathak, Li, Li, Xia, Neubig, Zhang, and Ramanan]{lin2024genaibench}
Z.~Lin, D.~Pathak, B.~Li, J.~Li, X.~Xia, G.~Neubig, P.~Zhang, and D.~Ramanan.
\newblock Evaluating text-to-visual generation with image-to-text generation.
\newblock \emph{arXiv preprint arXiv:2404.01291}, 2024.

\bibitem[Lipman et~al.(2023{\natexlab{a}})Lipman, Chen, Ben-Hamu, Nickel, and Le]{lipman2022flow}
Y.~Lipman, R.~T. Chen, H.~Ben-Hamu, M.~Nickel, and M.~Le.
\newblock Flow matching for generative modeling.
\newblock In \emph{International Conference on Machine Learning (ICLR)}, 2023{\natexlab{a}}.

\bibitem[Lipman et~al.(2023{\natexlab{b}})Lipman, Chen, Ben-Hamu, Nickel, and Le]{ipman2023flowmatching}
Y.~Lipman, R.~T.~Q. Chen, H.~Ben-Hamu, M.~Nickel, and M.~Le.
\newblock Flow matching for generative modeling.
\newblock \emph{arXiv preprint arXiv:2210.02747}, 2023{\natexlab{b}}.

\bibitem[Liu et~al.(2025)Liu, Han, Xing, Yin, Wang, Cheng, Liao, Wang, Fu, Han, et~al.]{liu2025step1x}
S.~Liu, Y.~Han, P.~Xing, F.~Yin, R.~Wang, W.~Cheng, J.~Liao, Y.~Wang, H.~Fu, C.~Han, et~al.
\newblock Step1x-edit: A practical framework for general image editing.
\newblock \emph{arXiv preprint arXiv:2504.17761}, 2025.

\bibitem[Loshchilov and Hutter(2019)]{loshchilov2019adamw}
I.~Loshchilov and F.~Hutter.
\newblock Decoupled weight decay regularization.
\newblock \emph{arXiv preprint arXiv:1711.05101}, 2019.

\bibitem[Lu et~al.(2022)Lu, Zhou, Bao, Chen, Li, and Zhu]{lu2022dpm_solver}
C.~Lu, Y.~Zhou, F.~Bao, J.~Chen, C.~Li, and J.~Zhu.
\newblock Dpm-solver: A fast ode solver for diffusion probabilistic model sampling in around 10 steps.
\newblock \emph{Advances in neural information processing systems}, 35:\penalty0 5775--5787, 2022.

\bibitem[Lu et~al.(2025)Lu, Zhou, Bao, Chen, Li, and Zhu]{lu2025dpm_plus}
C.~Lu, Y.~Zhou, F.~Bao, J.~Chen, C.~Li, and J.~Zhu.
\newblock Dpm-solver++: Fast solver for guided sampling of diffusion probabilistic models.
\newblock \emph{Machine Intelligence Research}, pages 1--22, 2025.

\bibitem[Ma et~al.(2025{\natexlab{a}})Ma, Huang, Yan, Chen, Duan, Yin, Wan, Ming, Song, Chen, Zhou, Sun, Zhou, Zhou, Tan, An, Chen, Ji, Wu, Sun, Han, Wei, Ge, Li, Wang, Huang, Wang, Li, Miao, Xu, Wu, Yu, Shi, Hu, Liu, Yu, Yang, Huang, Yan, Feng, Nie, Jia, Hu, Chen, Yan, Wang, Guo, Xiong, Xiong, Gong, Wu, Wu, Wu, Yang, Liu, Li, Zhang, Guo, Lin, Li, Liu, Xia, Zhao, Tan, Huang, Shi, Li, Li, Cheng, Wang, Chen, He, Liang, Sun, Sun, Wang, Pang, Yang, Liu, Liu, Gao, Cao, Wang, Ming, He, Zhao, Zhang, Zeng, Liu, Yang, Dai, Yu, Li, Deng, Wang, Wang, Lu, Chen, Luo, Luo, Yin, Feng, Yang, Tang, Zhang, Yang, Jiao, Chen, Li, Zhou, Zhang, Zhang, Zhu, Shum, and Jiang]{Step-Video-T2V}
G.~Ma, H.~Huang, K.~Yan, L.~Chen, N.~Duan, S.~Yin, C.~Wan, R.~Ming, X.~Song, X.~Chen, Y.~Zhou, D.~Sun, D.~Zhou, J.~Zhou, K.~Tan, K.~An, M.~Chen, W.~Ji, Q.~Wu, W.~Sun, X.~Han, Y.~Wei, Z.~Ge, A.~Li, B.~Wang, B.~Huang, B.~Wang, B.~Li, C.~Miao, C.~Xu, C.~Wu, C.~Yu, D.~Shi, D.~Hu, E.~Liu, G.~Yu, G.~Yang, G.~Huang, G.~Yan, H.~Feng, H.~Nie, H.~Jia, H.~Hu, H.~Chen, H.~Yan, H.~Wang, H.~Guo, H.~Xiong, H.~Xiong, J.~Gong, J.~Wu, J.~Wu, J.~Wu, J.~Yang, J.~Liu, J.~Li, J.~Zhang, J.~Guo, J.~Lin, K.~Li, L.~Liu, L.~Xia, L.~Zhao, L.~Tan, L.~Huang, L.~Shi, M.~Li, M.~Li, M.~Cheng, N.~Wang, Q.~Chen, Q.~He, Q.~Liang, Q.~Sun, R.~Sun, R.~Wang, S.~Pang, S.~Yang, S.~Liu, S.~Liu, S.~Gao, T.~Cao, T.~Wang, W.~Ming, W.~He, X.~Zhao, X.~Zhang, X.~Zeng, X.~Liu, X.~Yang, Y.~Dai, Y.~Yu, Y.~Li, Y.~Deng, Y.~Wang, Y.~Wang, Y.~Lu, Y.~Chen, Y.~Luo, Y.~Luo, Y.~Yin, Y.~Feng, Y.~Yang, Z.~Tang, Z.~Zhang, Z.~Yang, B.~Jiao, J.~Chen, J.~Li, S.~Zhou, X.~Zhang, X.~Zhang, Y.~Zhu, H.-Y. Shum, and D.~Jiang.
\newblock Step-video-t2v technical report: The practice, challenges, and future of video foundation model, 2025{\natexlab{a}}.
\newblock URL \url{https://arxiv.org/abs/2502.10248}.

\bibitem[Ma et~al.(2025{\natexlab{b}})Ma, Sun, Ma, Tang, Ma, Wang, Li, Dai, Shi, Ju, et~al.]{ma2025token-shuffle}
X.~Ma, P.~Sun, H.~Ma, H.~Tang, C.-Y. Ma, J.~Wang, K.~Li, X.~Dai, Y.~Shi, X.~Ju, et~al.
\newblock Token-shuffle: Towards high-resolution image generation with autoregressive models.
\newblock \emph{arXiv preprint arXiv:2504.17789}, 2025{\natexlab{b}}.

\bibitem[Meng et~al.(2023)Meng, Rombach, Gao, Kingma, Ermon, Ho, and Salimans]{meng2023distillation}
C.~Meng, R.~Rombach, R.~Gao, D.~Kingma, S.~Ermon, J.~Ho, and T.~Salimans.
\newblock On distillation of guided diffusion models.
\newblock In \emph{Proceedings of the IEEE/CVF conference on computer vision and pattern recognition}, pages 14297--14306, 2023.

\bibitem[Niu et~al.(2025)Niu, Ning, Zheng, Lin, Jin, Liao, Ning, Zhu, and Yuan]{niu2025wise}
Y.~Niu, M.~Ning, M.~Zheng, B.~Lin, P.~Jin, J.~Liao, K.~Ning, B.~Zhu, and L.~Yuan.
\newblock Wise: A world knowledge-informed semantic evaluation for text-to-image generation.
\newblock \emph{arXiv preprint arXiv:2503.07265}, 2025.

\bibitem[Oliveira and de~Matos(2025)]{oliveira2025storyreasoning}
D.~A. Oliveira and D.~M. de~Matos.
\newblock Storyreasoning dataset: Using chain-of-thought for scene understanding and grounded story generation.
\newblock \emph{arXiv preprint arXiv:2505.10292}, 2025.

\bibitem[OpenAI(2025{\natexlab{a}})]{gpt4.1}
OpenAI.
\newblock Introducing gpt-4.1 in the api.
\newblock \emph{OpenAI Blog}, 2025{\natexlab{a}}.
\newblock URL \url{https://openai.com/index/gpt-4-1}.

\bibitem[OpenAI(2025{\natexlab{b}})]{openai2024gpt4o_image}
OpenAI.
\newblock Introducing 4o image generation, 2025{\natexlab{b}}.
\newblock URL \url{https://openai.com/index/introducing-4o-image-generation}.

\bibitem[Pan et~al.(2025)Pan, Shukla, Singh, Zhao, Mishra, Wang, Xu, Chen, Li, Juefei-Xu, Hou, and Xie]{pan2025transfer}
X.~Pan, S.~N. Shukla, A.~Singh, Z.~Zhao, S.~K. Mishra, J.~Wang, Z.~Xu, J.~Chen, K.~Li, F.~Juefei-Xu, J.~Hou, and S.~Xie.
\newblock Transfer between modalities with metaqueries.
\newblock \emph{arXiv preprint arXiv:2504.06256}, 2025.

\bibitem[Peng et~al.(2024)Peng, Cui, Tang, Qi, Dong, Bai, Han, Ge, Zhang, and Xia]{peng2024dreambench++}
Y.~Peng, Y.~Cui, H.~Tang, Z.~Qi, R.~Dong, J.~Bai, C.~Han, Z.~Ge, X.~Zhang, and S.-T. Xia.
\newblock Dreambench++: A human-aligned benchmark for personalized image generation.
\newblock \emph{arXiv preprint arXiv:2406.16855}, 2024.

\bibitem[Podell et~al.(2024)Podell, English, Lacey, Blattmann, Dockhorn, M{\"u}ller, Penna, and Rombach]{sdxl}
D.~Podell, Z.~English, K.~Lacey, A.~Blattmann, T.~Dockhorn, J.~M{\"u}ller, J.~Penna, and R.~Rombach.
\newblock Sdxl: Improving latent diffusion models for high-resolution image synthesis.
\newblock In \emph{International Conference on Learning Representations (ICLR)}, 2024.

\bibitem[Qin et~al.(2025)Qin, Zhuo, Xin, Du, Li, Fu, Lu, Yuan, Li, Liu, et~al.]{qin2025lumina2}
Q.~Qin, L.~Zhuo, Y.~Xin, R.~Du, Z.~Li, B.~Fu, Y.~Lu, J.~Yuan, X.~Li, D.~Liu, et~al.
\newblock Lumina-image 2.0: A unified and efficient image generative framework.
\newblock \emph{arXiv preprint arXiv:2503.21758}, 2025.

\bibitem[Radford et~al.(2018)Radford, Narasimhan, Salimans, Sutskever, et~al.]{gpt1}
A.~Radford, K.~Narasimhan, T.~Salimans, I.~Sutskever, et~al.
\newblock Improving language understanding by generative pre-training.
\newblock \emph{San Francisco, CA, USA}, 2018.

\bibitem[Radford et~al.(2019)Radford, Wu, Child, Luan, Amodei, Sutskever, et~al.]{gpt2}
A.~Radford, J.~Wu, R.~Child, D.~Luan, D.~Amodei, I.~Sutskever, et~al.
\newblock Language models are unsupervised multitask learners.
\newblock \emph{OpenAI blog}, 2019.

\bibitem[Rafailov et~al.(2023)Rafailov, Sharma, Mitchell, Manning, Ermon, and Finn]{rafailov2023direct}
R.~Rafailov, A.~Sharma, E.~Mitchell, C.~D. Manning, S.~Ermon, and C.~Finn.
\newblock Direct preference optimization: Your language model is secretly a reward model.
\newblock \emph{Advances in neural information processing systems (NeurIPS)}, 2023.

\bibitem[Rafailov et~al.(2024)Rafailov, Sharma, Mitchell, Ermon, Manning, and Finn]{rafailov2024dpo}
R.~Rafailov, A.~Sharma, E.~Mitchell, S.~Ermon, C.~D. Manning, and C.~Finn.
\newblock Direct preference optimization: Your language model is secretly a reward model.
\newblock \emph{arXiv preprint arXiv:2305.18290}, 2024.

\bibitem[Rombach et~al.(2022)Rombach, Blattmann, Lorenz, Esser, and Ommer]{sdv15}
R.~Rombach, A.~Blattmann, D.~Lorenz, P.~Esser, and B.~Ommer.
\newblock High-resolution image synthesis with latent diffusion models.
\newblock In \emph{Proceedings of the IEEE/CVF Conference on Computer Vision and Pattern Recognition (CVPR)}, 2022.

\bibitem[Shi et~al.(2024)Shi, Wang, and Huang]{shi2024seededit}
Y.~Shi, P.~Wang, and W.~Huang.
\newblock Seededit: Align image re-generation to image editing.
\newblock \emph{arXiv preprint arXiv:2411.06686}, 2024.

\bibitem[Stability-AI(2024)]{2024sd3.5}
Stability-AI.
\newblock stable-diffusion-3.5-large, 2024.
\newblock URL \url{https://github.com/Stability-AI/sd3.5}.

\bibitem[Su et~al.(2024)Su, Ahmed, Lu, Pan, Bo, and Liu]{su2024roformer}
J.~Su, M.~Ahmed, Y.~Lu, S.~Pan, W.~Bo, and Y.~Liu.
\newblock Roformer: Enhanced transformer with rotary position embedding.
\newblock \emph{Neurocomputing}, 2024.

\bibitem[Sun et~al.(2024{\natexlab{a}})Sun, Jiang, Chen, Zhang, Peng, Luo, and Yuan]{sun2024llamagen}
P.~Sun, Y.~Jiang, S.~Chen, S.~Zhang, B.~Peng, P.~Luo, and Z.~Yuan.
\newblock Autoregressive model beats diffusion: Llama for scalable image generation.
\newblock \emph{arXiv preprint arXiv:2406.06525}, 2024{\natexlab{a}}.

\bibitem[Sun et~al.(2023)Sun, Yu, Cui, Zhang, Zhang, Wang, Gao, Liu, Huang, and Wang]{emu1}
Q.~Sun, Q.~Yu, Y.~Cui, F.~Zhang, X.~Zhang, Y.~Wang, H.~Gao, J.~Liu, T.~Huang, and X.~Wang.
\newblock Emu: Generative pretraining in multimodality.
\newblock In \emph{International Conference on Learning Representations (ICLR)}, 2023.

\bibitem[Sun et~al.(2024{\natexlab{b}})Sun, Cui, Zhang, Zhang, Yu, Wang, Rao, Liu, Huang, and Wang]{emu2}
Q.~Sun, Y.~Cui, X.~Zhang, F.~Zhang, Q.~Yu, Y.~Wang, Y.~Rao, J.~Liu, T.~Huang, and X.~Wang.
\newblock Generative multimodal models are in-context learners.
\newblock In \emph{Proceedings of the IEEE/CVF Conference on Computer Vision and Pattern Recognition (CVPR)}, 2024{\natexlab{b}}.

\bibitem[Sun et~al.(2024{\natexlab{c}})Sun, Bao, Wang, Peng, Dong, Huang, Wang, and Wei]{sun2024latentlm}
Y.~Sun, H.~Bao, W.~Wang, Z.~Peng, L.~Dong, S.~Huang, J.~Wang, and F.~Wei.
\newblock Multimodal latent language modeling with next-token diffusion.
\newblock \emph{arXiv preprint arXiv:2412.08635}, 2024{\natexlab{c}}.

\bibitem[team(2025)]{2025Kolors2}
K.~K. team.
\newblock Kolors2.0, 2025.
\newblock URL \url{https://app.klingai.com/cn}.

\bibitem[team(2024)]{2024recraftv3}
R.~team.
\newblock Recraft v3, 2024.
\newblock URL \url{https://www.recraft.ai/blog/recraft-introduces-a-revolutionary-ai-model-that-thinks-in-design-language}.

\bibitem[Tian et~al.(2024)Tian, Jiang, Yuan, Peng, and Wang]{tian2024visual}
K.~Tian, Y.~Jiang, Z.~Yuan, B.~Peng, and L.~Wang.
\newblock Visual autoregressive modeling: Scalable image generation via next-scale prediction.
\newblock \emph{Advances in neural information processing systems (NeurIPS)}, 2024.

\bibitem[Tong et~al.(2024)Tong, Fan, Zhu, Xiong, Chen, Sinha, Rabbat, LeCun, Xie, and Liu]{tong2024metamorph}
S.~Tong, D.~Fan, J.~Zhu, Y.~Xiong, X.~Chen, K.~Sinha, M.~Rabbat, Y.~LeCun, S.~Xie, and Z.~Liu.
\newblock Metamorph: Multimodal understanding and generation via instruction tuning.
\newblock \emph{arXiv preprint arXiv:2412.14164}, 2024.

\bibitem[Tschannen et~al.(2024)Tschannen, Pinto, and Kolesnikov]{tschannen2024jetformer}
M.~Tschannen, A.~S. Pinto, and A.~Kolesnikov.
\newblock Jetformer: An autoregressive generative model of raw images and text.
\newblock \emph{arXiv preprint arXiv:2411.19722}, 2024.

\bibitem[Tschannen et~al.(2025)Tschannen, Eastwood, and Mentzer]{tschannen2025givt}
M.~Tschannen, C.~Eastwood, and F.~Mentzer.
\newblock Givt: Generative infinite-vocabulary transformers.
\newblock In \emph{European Conference on Computer Vision (ECCV)}, 2025.

\bibitem[Wallace et~al.(2024)Wallace, Dang, Rafailov, Zhou, Lou, Purushwalkam, Ermon, Xiong, Joty, and Naik]{wallace2024diffusiondpo}
B.~Wallace, M.~Dang, R.~Rafailov, L.~Zhou, A.~Lou, S.~Purushwalkam, S.~Ermon, C.~Xiong, S.~Joty, and N.~Naik.
\newblock Diffusion model alignment using direct preference optimization.
\newblock \emph{Proceedings of the IEEE/CVF Conference on Computer Vision and Pattern Recognition (CVPR)}, 2024.

\bibitem[Wang et~al.(2025{\natexlab{a}})Wang, Wang, Wan, Huang, Hu, Jia, Nie, Li, Chen, Chen, et~al.]{stepfun2025step3largeaffordablemodelsystem}
B.~Wang, B.~Wang, C.~Wan, G.~Huang, H.~Hu, H.~Jia, H.~Nie, M.~Li, N.~Chen, S.~Chen, et~al.
\newblock Step-3 is large yet affordable: Model-system co-design for cost-effective decoding.
\newblock \emph{arXiv preprint arXiv:2507.19427}, 2025{\natexlab{a}}.

\bibitem[Wang et~al.(2025{\natexlab{b}})Wang, Yue, Zhang, Chen, Bi, Zhang, Song, Chan, Pan, Wu, et~al.]{wang2025selftok}
B.~Wang, Z.~Yue, F.~Zhang, S.~Chen, L.~Bi, J.~Zhang, X.~Song, K.~Y. Chan, J.~Pan, W.~Wu, et~al.
\newblock Selftok: Discrete visual tokens of autoregression, by diffusion, and for reasoning.
\newblock \emph{arXiv preprint arXiv:2505.07538}, 2025{\natexlab{b}}.

\bibitem[Wang et~al.(2025{\natexlab{c}})Wang, Tian, Wang, Zhang, Huang, Wu, and Jiang]{wang2025simplear}
J.~Wang, Z.~Tian, X.~Wang, X.~Zhang, W.~Huang, Z.~Wu, and Y.-G. Jiang.
\newblock Simplear: Pushing the frontier of autoregressive visual generation through pretraining, sft, and rl.
\newblock \emph{arXiv preprint arXiv:2504.11455}, 2025{\natexlab{c}}.

\bibitem[Wang et~al.(2024{\natexlab{a}})Wang, Bai, Tan, Wang, Fan, Bai, Chen, Liu, Wang, Ge, Fan, Dang, Du, Ren, Men, Liu, Zhou, Zhou, and Lin]{qwen2vl}
P.~Wang, S.~Bai, S.~Tan, S.~Wang, Z.~Fan, J.~Bai, K.~Chen, X.~Liu, J.~Wang, W.~Ge, Y.~Fan, K.~Dang, M.~Du, X.~Ren, R.~Men, D.~Liu, C.~Zhou, J.~Zhou, and J.~Lin.
\newblock Qwen2-vl: Enhancing vision-language model's perception of the world at any resolution.
\newblock \emph{arXiv preprint arXiv:2409.12191}, 2024{\natexlab{a}}.

\bibitem[Wang et~al.(2024{\natexlab{b}})Wang, Zhang, Luo, Sun, Cui, Wang, Zhang, Wang, Li, Yu, et~al.]{emu3}
X.~Wang, X.~Zhang, Z.~Luo, Q.~Sun, Y.~Cui, J.~Wang, F.~Zhang, Y.~Wang, Z.~Li, Q.~Yu, et~al.
\newblock Emu3: Next-token prediction is all you need.
\newblock \emph{arXiv preprint arxiv:2409.18869}, 2024{\natexlab{b}}.

\bibitem[Wang et~al.(2025{\natexlab{d}})Wang, Yang, Zhao, Zhang, Liu, Zhou, and Xie]{wang2025gpt-edit-1_5M}
Y.~Wang, S.~Yang, B.~Zhao, L.~Zhang, Q.~Liu, Y.~Zhou, and C.~Xie.
\newblock Gpt-image-edit-1.5 m: A million-scale, gpt-generated image dataset.
\newblock \emph{arXiv preprint arXiv:2507.21033}, 2025{\natexlab{d}}.

\bibitem[Wei et~al.(2022)Wei, Wang, Schuurmans, Bosma, Xia, Chi, Le, Zhou, et~al.]{COT}
J.~Wei, X.~Wang, D.~Schuurmans, M.~Bosma, F.~Xia, E.~Chi, Q.~V. Le, D.~Zhou, et~al.
\newblock Chain-of-thought prompting elicits reasoning in large language models.
\newblock \emph{Advances in neural information processing systems (NeurIPS)}, 2022.

\bibitem[Wu et~al.(2025{\natexlab{a}})Wu, Li, Zhou, Lin, Gao, Yan, Yin, Bai, Xu, Chen, et~al.]{wu2025qwen_image}
C.~Wu, J.~Li, J.~Zhou, J.~Lin, K.~Gao, K.~Yan, S.-m. Yin, S.~Bai, X.~Xu, Y.~Chen, et~al.
\newblock Qwen-image technical report.
\newblock \emph{arXiv preprint arXiv:2508.02324}, 2025{\natexlab{a}}.

\bibitem[Wu et~al.(2025{\natexlab{b}})Wu, Zheng, Yan, Xiao, Luo, Wang, Li, Jiang, Liu, Zhou, et~al.]{wu2025omnigen2}
C.~Wu, P.~Zheng, R.~Yan, S.~Xiao, X.~Luo, Y.~Wang, W.~Li, X.~Jiang, Y.~Liu, J.~Zhou, et~al.
\newblock Omnigen2: Exploration to advanced multimodal generation.
\newblock \emph{arXiv preprint arXiv:2506.18871}, 2025{\natexlab{b}}.

\bibitem[Wu et~al.(2024)Wu, Zhang, Chen, Tang, Li, Fang, Zhu, Xie, Yin, Yi, et~al.]{vila-u}
Y.~Wu, Z.~Zhang, J.~Chen, H.~Tang, D.~Li, Y.~Fang, L.~Zhu, E.~Xie, H.~Yin, L.~Yi, et~al.
\newblock Vila-u: a unified foundation model integrating visual understanding and generation.
\newblock \emph{arXiv preprint arXiv:2409.04429}, 2024.

\bibitem[Xiao et~al.(2024)Xiao, Wang, Zhou, Yuan, Xing, Yan, Li, Wang, Huang, and Liu]{xiao2024omnigen}
S.~Xiao, Y.~Wang, J.~Zhou, H.~Yuan, X.~Xing, R.~Yan, C.~Li, S.~Wang, T.~Huang, and Z.~Liu.
\newblock Omnigen: Unified image generation.
\newblock \emph{arXiv preprint arXiv:2409.11340}, 2024.

\bibitem[Xie et~al.(2025{\natexlab{a}})Xie, Chen, Zhao, Yu, Zhu, Wu, Lin, Zhang, Li, Chen, et~al.]{xie2025sana}
E.~Xie, J.~Chen, Y.~Zhao, J.~Yu, L.~Zhu, C.~Wu, Y.~Lin, Z.~Zhang, M.~Li, J.~Chen, et~al.
\newblock Sana 1.5: Efficient scaling of training-time and inference-time compute in linear diffusion transformer.
\newblock \emph{arXiv preprint arXiv:2501.18427}, 2025{\natexlab{a}}.

\bibitem[Xie et~al.(2024)Xie, Mao, Bai, Zhang, Wang, Lin, Gu, Chen, Yang, and Shou]{show-o}
J.~Xie, W.~Mao, Z.~Bai, D.~J. Zhang, W.~Wang, K.~Q. Lin, Y.~Gu, Z.~Chen, Z.~Yang, and M.~Z. Shou.
\newblock Show-o: One single transformer to unify multimodal understanding and generation.
\newblock \emph{arXiv preprint arxiv:2408.12528}, 2024.

\bibitem[Xie et~al.(2025{\natexlab{b}})Xie, Yang, and Shou]{xie2025showo2}
J.~Xie, Z.~Yang, and M.~Z. Shou.
\newblock Show-o2: Improved native unified multimodal models.
\newblock \emph{arXiv preprint arXiv:2506.15564}, 2025{\natexlab{b}}.

\bibitem[Xu et~al.(2023)Xu, Liu, Wu, Tong, Li, Ding, Tang, and Dong]{xu2023imagereward}
J.~Xu, X.~Liu, Y.~Wu, Y.~Tong, Q.~Li, M.~Ding, J.~Tang, and Y.~Dong.
\newblock Imagereward: Learning and evaluating human preferences for text-to-image generation.
\newblock \emph{Advances in Neural Information Processing Systems (NeurIPS)}, 2023.

\bibitem[Yang et~al.(2024)Yang, Yang, Zhang, Hui, Zheng, Yu, Li, Liu, Huang, Wei, Lin, Yang, Tu, Zhang, Yang, Yang, Zhou, Lin, Dang, Lu, Bao, Yang, Yu, Li, Xue, Zhang, Zhu, Men, Lin, Li, Xia, Ren, Ren, Fan, Su, Zhang, Wan, Liu, Cui, Zhang, and Qiu]{qwen2.5}
A.~Yang, B.~Yang, B.~Zhang, B.~Hui, B.~Zheng, B.~Yu, C.~Li, D.~Liu, F.~Huang, H.~Wei, H.~Lin, J.~Yang, J.~Tu, J.~Zhang, J.~Yang, J.~Yang, J.~Zhou, J.~Lin, K.~Dang, K.~Lu, K.~Bao, K.~Yang, L.~Yu, M.~Li, M.~Xue, P.~Zhang, Q.~Zhu, R.~Men, R.~Lin, T.~Li, T.~Xia, X.~Ren, X.~Ren, Y.~Fan, Y.~Su, Y.~Zhang, Y.~Wan, Y.~Liu, Z.~Cui, Z.~Zhang, and Z.~Qiu.
\newblock Qwen2.5 technical report.
\newblock \emph{arXiv preprint arXiv:2412.15115}, 2024.

\bibitem[Yang et~al.(2025)Yang, Li, Fan, Tian, and Wang]{yang2025latent_make_good_tokenizer}
J.~Yang, T.~Li, L.~Fan, Y.~Tian, and Y.~Wang.
\newblock Latent denoising makes good visual tokenizers.
\newblock \emph{arXiv preprint arXiv:2507.15856}, 2025.

\bibitem[Yao et~al.(2025)Yao, Yang, and Wang]{yao2025vavae}
J.~Yao, B.~Yang, and X.~Wang.
\newblock Reconstruction vs. generation: Taming optimization dilemma in latent diffusion models.
\newblock In \emph{Proceedings of the Computer Vision and Pattern Recognition Conference (CVPR)}, 2025.

\bibitem[Ye et~al.(2025)Ye, He, Li, Lin, Yuan, Yan, Hou, and Yuan]{ye2025imgedit}
Y.~Ye, X.~He, Z.~Li, B.~Lin, S.~Yuan, Z.~Yan, B.~Hou, and L.~Yuan.
\newblock Imgedit: A unified image editing dataset and benchmark.
\newblock \emph{arXiv preprint arXiv:2505.20275}, 2025.

\bibitem[Yu et~al.(2022)Yu, Xu, Koh, Luong, Baid, Wang, Vasudevan, Ku, Yang, Ayan, Hutchinson, Han, Parekh, Li, Zhang, Baldridge, and Wu]{parti}
J.~Yu, Y.~Xu, J.~Y. Koh, T.~Luong, G.~Baid, Z.~Wang, V.~Vasudevan, A.~Ku, Y.~Yang, B.~K. Ayan, B.~Hutchinson, W.~Han, Z.~Parekh, X.~Li, H.~Zhang, J.~Baldridge, and Y.~Wu.
\newblock Scaling autoregressive models for content-rich text-to-image generation.
\newblock In \emph{TMLR}, 2022.

\bibitem[Yu et~al.(2023)Yu, Lezama, Gundavarapu, Versari, Sohn, Minnen, Cheng, Birodkar, Gupta, Gu, et~al.]{yu2023magvitv2}
L.~Yu, J.~Lezama, N.~B. Gundavarapu, L.~Versari, K.~Sohn, D.~Minnen, Y.~Cheng, V.~Birodkar, A.~Gupta, X.~Gu, et~al.
\newblock Language model beats diffusion--tokenizer is key to visual generation.
\newblock \emph{arXiv preprint arXiv:2310.05737}, 2023.

\bibitem[Yu et~al.(2024{\natexlab{a}})Yu, Chow, Yue, Pan, Wu, Wan, Li, Tang, Zhang, and Zhuang]{yu2024anyedit}
Q.~Yu, W.~Chow, Z.~Yue, K.~Pan, Y.~Wu, X.~Wan, J.~Li, S.~Tang, H.~Zhang, and Y.~Zhuang.
\newblock Anyedit: Mastering unified high-quality image editing for any idea.
\newblock \emph{arXiv preprint arXiv:2411.15738}, 2024{\natexlab{a}}.

\bibitem[Yu et~al.(2024{\natexlab{b}})Yu, Weber, Deng, Shen, Cremers, and Chen]{yu2024titok}
Q.~Yu, M.~Weber, X.~Deng, X.~Shen, D.~Cremers, and L.-C. Chen.
\newblock An image is worth 32 tokens for reconstruction and generation.
\newblock \emph{Advances in Neural Information Processing Systems (NeurIPS)}, 2024{\natexlab{b}}.

\bibitem[Z.ai(2025)]{2025cogview4}
T.~Z.ai.
\newblock Cogview4, 2025.
\newblock URL \url{https://github.com/THUDM/CogView4}.

\bibitem[Zhang et~al.(2023{\natexlab{a}})Zhang, Mo, Chen, Sun, and Su]{zhang2023magicbrush}
K.~Zhang, L.~Mo, W.~Chen, H.~Sun, and Y.~Su.
\newblock Magicbrush: A manually annotated dataset for instruction-guided image editing.
\newblock In \emph{Advances in neural information processing systems (NeurIPS)}, 2023{\natexlab{a}}.

\bibitem[Zhang et~al.(2023{\natexlab{b}})Zhang, Rao, and Agrawala]{zhang2023controlnet}
L.~Zhang, A.~Rao, and M.~Agrawala.
\newblock Adding conditional control to text-to-image diffusion models.
\newblock In \emph{Proceedings of the IEEE/CVF international conference on computer vision (ICCV)}, 2023{\natexlab{b}}.

\bibitem[Zhang et~al.(2025)Zhang, Zhang, Li, Sun, Shen, Lu, Zhao, Zhuang, and Bing]{mmtextbook}
W.~Zhang, H.~Zhang, X.~Li, J.~Sun, Y.~Shen, W.~Lu, D.~Zhao, Y.~Zhuang, and L.~Bing.
\newblock 2.5 years in class: A multimodal textbook for vision-language pretraining.
\newblock \emph{arXiv preprint arXiv:2501.00958}, 2025.

\bibitem[Zheng et~al.(2022)Zheng, Vuong, Cai, and Phung]{movqgan}
C.~Zheng, T.-L. Vuong, J.~Cai, and D.~Phung.
\newblock Movq: Modulating quantized vectors for high-fidelity image generation.
\newblock \emph{Advances in Neural Information Processing Systems (NeurIPS)}, 2022.

\bibitem[Zhou et~al.(2025)Zhou, Yu, Babu, Tirumala, Yasunaga, Shamis, Kahn, Ma, Zettlemoyer, and Levy]{transfusion}
C.~Zhou, L.~Yu, A.~Babu, K.~Tirumala, M.~Yasunaga, L.~Shamis, J.~Kahn, X.~Ma, L.~Zettlemoyer, and O.~Levy.
\newblock Transfusion: Predict the next token and diffuse images with one multi-modal model.
\newblock \emph{International Conference on Learning Representations (ICLR)}, 2025.

\end{thebibliography}

\setcounter{figure}{0}
\makeatletter
\renewcommand{\thefigure}{A\@arabic\c@figure}
\makeatother

\setcounter{table}{0}
\makeatletter
\renewcommand{\thetable}{A\@arabic\c@table}
\makeatother

\end{document}